\documentclass[10pt,twocolumn,letterpaper]{article}

\usepackage{cvpr}
\usepackage{times}
\usepackage{epsfig}
\usepackage{graphicx}
\usepackage{amsmath}
\usepackage{amssymb}
\usepackage{booktabs,multirow,url,bm,appendix}       
\usepackage[breaklinks=true,letterpaper=true,colorlinks,bookmarks=false]{hyperref}

\cvprfinalcopy 

\def\cvprPaperID{****} 

\begin{document}

\title{RefineDetLite: A Lightweight One-stage Object Detection Framework for CPU-only Devices}

\author{Chen Chen$^1$, Mengyuan Liu$^1$, Xiandong Meng$^2$, Wanpeng Xiao$^1$, Qi Ju$^1$\\
$^1$Tencent TEG AI\quad$^2$The Hong Kong University of Science and Technology\\
{\tt\small \{beckhamchen, mengyuanliu, wanpengxiao, damonju\}@tencent.com, \tt\small xmengab@connect.ust.hk}
}

\maketitle

\begin{abstract}
Previous state-of-the-art real-time object detectors have been reported on GPUs which are extremely expensive for processing massive data and in resource-restricted scenarios.
Therefore, high efficiency object detectors on CPU-only devices are urgently-needed in industry. The floating-point operations (FLOPs\footnote{Here, FLOPs means the number of multiply-adds following~\cite{ShuffleNetV2}.}) of networks are not strictly proportional to the running speed on CPU devices, which inspires the design of an exactly ``fast'' and ``accurate'' object detector.  After investigating the concern gaps between classification networks and detection backbones, and following the design principles of efficient networks, we propose a lightweight residual-like backbone with large receptive fields and wide dimensions for low-level features, which are crucial for detection tasks. Correspondingly, we also design a light-head detection part to match the backbone capability. Furthermore, by analyzing the drawbacks of current one-stage detector training strategies, we also propose three orthogonal training strategies---IOU-guided loss, classes-aware weighting method and balanced multi-task training approach. Without bells and whistles, our proposed RefineDetLite achieves 26.8 mAP on the MSCOCO benchmark at a speed of 130 ms/pic on a single-thread CPU. The detection accuracy can be further increased to 29.6 mAP by integrating all the proposed training strategies, without apparent speed drop.
\end{abstract}

\section{Introduction}

Object detection is a fundamental technology in the computer vision society and is also a crucial component for many high-level artificial intelligence tasks, \eg, object tracking~\cite{TrackingDetection}, vision-language transferring~\cite{ImageCaptionBottomUp,ImageCaptionGAN}, surveillance, autonomous driving~\cite{SqueezeDet} and robotics. Benefited from the rapid development of deep learning, the accuracy of object detection has been greatly improved. However, with the explosive growth of social media information, the high computational complexity seriously hinders the wide applications of object detection algorithms. Therefore, much attention has been paid to the study of how to make trade-off between detection accuracy and implementation complexity. Thanks to the powerful parallel processing ability of GPUs, many researchers claimed they have achieved real-time detection. However, GPUs are still extremely high cost in terms of dealing with massive data. Consequently, research into fast object detection pipelines on computationally constrained devices (\eg, CPU-only computers and mobile devices) is extremely urgent.

Inspired by the pioneering deep-learning-based R-CNN serials (\cite{RCNN,FastRCNN,FasterRCNN}), most state-of-the-art detectors are inclined to exploit classical classification networks~\cite{ResNet,VGG} as the backbone part. Obviously, the computational complexity of backbone networks is the important bottleneck that affects the running efficiency of the whole detector, and hence many lightweight algorithms employ famous efficient convolution networks~\cite{MobileNet,MobileNetV2,MobileNetV3,ShuffleNet,ShuffleNetV2,SqueezeNet,SqueezeNext}) instead. However, as pointed out in~\cite{DetNet}, there exists gaps between the design principles of classification and detection networks. For instance, the larger receptive fields and wider feature vectors of early stages are crucial for improving localization ability, while classification networks care only about the feature representation ability of the last layer. Therefore, directly employing classification networks as the backbones maybe not the optimal strategy. Additionally, another important issue that must be recognized is that the number of FLOPs is not strictly proportional to the running time since many other factors (\eg, memory access cost and degree of parallelism) impact the practical network latency~\cite{ShuffleNetV2}. Therefore, how to design an actually ``\textbf{fast}'' detection backbone network running on a CPU is a critical demand in industrial practice.

Typically, CNN-based object detectors are categorized into either two-stage detectors or one-stage detectors based on different processings of the detection part. Two-stage~\cite{FasterRCNN} detectors usually contain a region proposal network (RPN), a RoI warping module and a localization and classification subnet. More elegantly, one-stage detectors~\cite{YOLO,YOLOV2,YOLOV3,SSD} directly output bounding boxes and classification probabilities through only once network forward pass. In general, two-stage detectors are usually considered to be more accurate on detection because of the bounding boxes ``\textbf{refinement}'' operation during the second stage, but are more time-consuming as compared to one-stage detectors. Intuitively, in the past few years, the majority of researchers have been dedicated to studying lightweight detection structures of one-stage detectors~\cite{urlYOLO,TinySSD,TinyDSOD,MobileNetV2,SqueezeDet}. But, the low detection accuracy cannot satisfy the practical requirements because of the coarse localization and classification through only single stage prediction. Therefore, other works were encouraged to develop or even automatically search lightweight detection architectures based on two-stage detectors~\cite{LightHead-RCNN,ThunderNet,NAS-FPN}, which have achieved a relatively higher detection accuracy under low computational complexity.

To inherit the merits of both two-stage (high accuracy) and one-stage (high efficiency) detectors, Zhang~\etal proposed a single-stage refinement network (RefineDet~\cite{RefineDet}), which can be viewed as a pseudo two-stage detector. By adding a lightweight FPN-like feature refinement stage, RefineDet simulates the anchor refining process during the second stage of the two-stage detector. Since RefineDet optimizes the anchor refinement loss and the object detection loss simultaneously, it can achieve high detection accuracy and efficiency at the same time. Inspired by this effective architecture and following the design principles of fast convolution networks suitable for detection, we propose a lightweight version of RefineDet, named RefineDetLite.

Besides the network architecture design, we further analyze the drawbacks of current one-stage detectors. With the aim of improving the detection accuracy without increasing any inference computational load, we concentrate on optimizing the loss functions and training strategies. First, we focus on the training data imbalance problem among different classes. Second, we propose an intersection-over-union (IOU)-guided loss to overcome the inconsistency between localization and classification confidences during training. Finally, going deep into the essence of object detection, we classify it as a multi-task training problem. In order to keep the balance of localization and detection losses during training, we introduce some trainable balancing coefficients and derive the final loss formulation in detail following the theoretical guidance of~\cite{MultiTask}.

In summary, the main contributions of our proposed efficient object detection framework are as follows:
\begin{itemize}
  \item Based on the RefineDet, analyzing the design key points of efficient convolutional networks, we propose a lightweight backbone and detection head specifically designed for the detection task. We call the entire network structure RefineDetLite.
  \item Without introducing any extra computational cost during inference, we propose some general training strategies (IOU-guided loss, classes-weighted loss and balanced multi-task training) to further improve the detection accuracy.
  \item RefineDetLite surpasses many state-of-the-art lightweight one-stage and two-stage detectors with faster running speed on CPUs and higher detection accuracy on the MSCOCO benchmark. It can achieve \textbf{29.6 mAP} on MSCOCO online \verb"test-dev" at a running speed of \textbf{131 ms/pic} on a single thread CPU\footnote{The CPU type is Intel i7-6700@3.40GHz}.
\end{itemize}

\section{Related Works}
\begin{figure*}[t]
  \centering
  \centerline{\includegraphics[width=16cm]{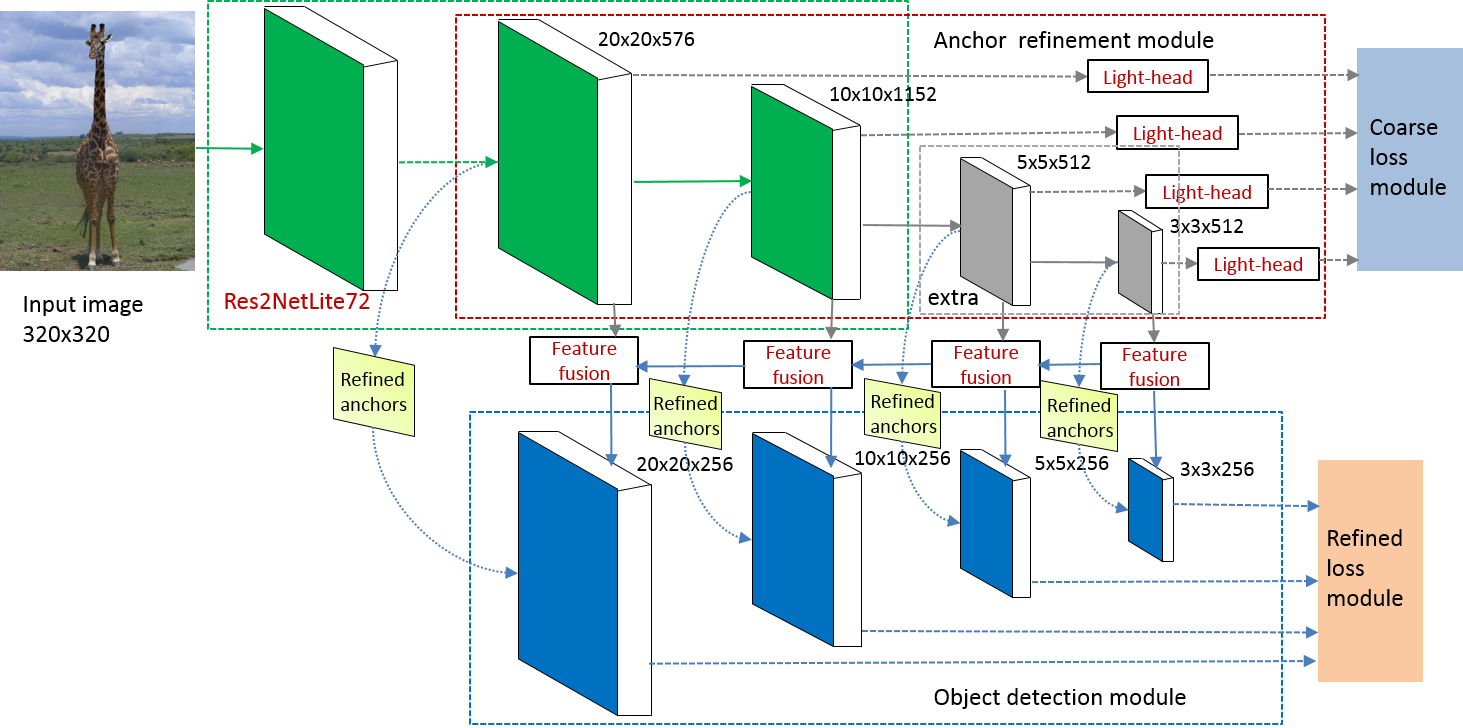}}
\caption{The overall framework of our proposed RefineDetLite.}
\label{fig:overall_framework}
\end{figure*}
\subsection{Deep-learning-based object detectors}
The well known R-CNN~\cite{RCNN} is the pioneer of the deep-learning-based object detectors, which creatively utilizes convolutional networks to predict object regions and class labels based on a sparse set of pre-extracted candidate region proposals. Encouraged by the R-CNN prototype, numerous successors boosted the performance of CNN-based detectors by optimizing candidate extracting approaches~\cite{FastRCNN,FasterRCNN,HyperNet}, feature fusion methods~\cite{FPN,MSCNN,TDM}, training strategies~\cite{GCNN,OHEM,A-fast-rcnn} and contextual reasoning~\cite{ION,ContextRef,RelationNet}.

In addition to the complex two-stage diagram, another fast-growing pipeline is the one-stage detectors, which directly predict object bounding boxes and category probabilities through a single forward pass and end-to-end training. Considered to be more straightforward and efficient, in the past few years, following YOLO~\cite{YOLO,YOLOV2,YOLOV3} and SSD~\cite{SSD}, a large number of researches have paid more attention to bridging the detection accuracy gap between two-stage and one-stage detectors. DSSD~\cite{DSSD}, FSSD~\cite{FSSD} and DSOD~\cite{DSOD} exploit different feature fusion methods to ameliorate the weak representation ability of low-level feature. Taking a step further, RefineDet~\cite{RefineDet} introduces an extra loss refinement stage to considerably improve small-object detection accuracy without significantly increasing the complexity.

\subsection{Efficient object detection}
Although many stat-of-the-art object detectors claim that they have achieved real-time detection speed, most of them were experimented on GPUs, which are still a heavy burden for personal users or industrial massive-data scenarios. Therefore, research on lightweight detection frameworks are prevalent in the object detection community. Noting the advantages described above, ideas about how to simplify one-stage detection architectures are overwhelming. SSDLite~\cite{MobileNetV2}, Tiny-SSD~\cite{Tiny-SSD} and Tiny-YOLO~\cite{urlYOLO} intuitively employ lightweight backbones and detection heads to replace original components of YOLO or SSD. To improve such naive thoughts, Tiny-DSOD~\cite{Tiny-DSOD} and Pelee~\cite{PeleeNet} propose more effective network simplification schemes based on advanced versions of SSD. However, even achieved speed boost, they perform poorly in detection accuracy. Alternatively, another group of researchers attempted to improve the detection efficiency by re-designing network structures manually~\cite{ThunderNet} or through neural architecture searching (NAS)~\cite{NAS-FPN} based on two-stage paradigms.

\subsection{Training and inference strategy optimization}

In addition to making great efforts to optimize the network structure, many researchers have been dedicated to distilling some general training or inference strategies to further improve the detection accuracy for existing detectors, with little increasing of computational complexity.

Inheriting the standard non-maximum suppression (NMS), soft-nms~\cite{soft-nms} slightly modifies this process to achieve a remarkable mAP increase. Taking this a step further, softer-nms~\cite{softer-nms} adds a small portion of parameters during training and inference by taking into account the uncertainty of bounding box regression. Goldman \etal~\cite{ DenselyDetection} further proposed a soft IOU layer and EM-Merger unit to reduce the bounding box prediction uncertainty.

To alleviate the well-known imbalance problem between positive and negative samples for one-stage detectors, online hard example mining (OHEM~\cite{OHEM}), focal loss~\cite{FocalLoss} and the gradient harmonizing mechanism (GHM) have been successively proposed.

Additionally, Yu \etal~\cite{Unitbox} noticed a long-time neglected misalignment problem---minimizing bounding box offsets does not strictly equal to maximizing IOU, the evaluation metric for regression accuracy. Hence, an IOU loss was invented to be directly used as a regression loss. Successively, GIOU~\cite{GIOU}, DIOU and CIOU~\cite{DIOU} were proposed to further improve the IOU-based evaluation metrics.
Another easily overlooked misalignment phenomenon for one-stage detectors is the inconsistency between the training hypothesis and inference configurations. Kong \etal~\cite{ConsistentOpt} proposed a consistent optimization strategy to bridge the gap.

\section{RefineDetLite}
In this section, we present in detail the RefineDetLite network architecture. Following the design principles of efficient convolutional networks on CPUs, our proposed detection network keeps the balance between efficiency and accuracy to the maximum degree.

\subsection{Overall framework}
Fig.~\ref{fig:overall_framework} elaborately illustrates the overall framework of RefineDetLite, which inherits the design idea of RefineDet~\cite{RefineDet}. It first extracts pyramidal features to predict coarse bounding boxes and decides whether the anchor is foreground or background, which is called the anchor refinement module (ARM). Then, the skillfully contrived object detection module (ODM) fuses the pyramidal features reversely and employs the refined anchors to further predict precise bounding boxes and exact object classes. Since the network outputs box regions and class probabilities through only one forward pass, it is still a one-stage object detector. However, the ingenious structure achieves high efficiency and detection accuracy simultaneously.

As demonstrated in~\cite{ThunderNet}, the input resolution should match the capability of the backbone network, so we fixed the input resolution as $320\times 320$ for the high efficiency backbone. Then, two groups of pyramidal features with the resolutions \{$20\times 20, 10\times 10, 5\times 5, 3\times 3$\} are used to calculate coarse losses and refined losses, respectively.

Since the original RefineDet is not concerned about the network efficiency on CPUs, we are enlightened to improve this structure by introducing a new efficient backbone, re-designing a light-head detection part and a lightweight feature fusion module.
\subsection{Backbone}
\begin{table}[t]
\centering
\caption{Overview of Res2NetLite architecture} \label{tab:Res2NetLite}
\resizebox{.99\columnwidth}{!}{
\begin{tabular}{c c|c|c}
\toprule[2pt]
\multicolumn{2}{c|}{\textbf{Stage}} & \textbf{Layer} & \textbf{Output Size} \\
\toprule[1pt]
\multicolumn{3}{c|}{input}                                           & 224$\times$224$\times$3 \\
\hline
\multirow{3}{*}{\textbf{Stage0}} & BatchNorm      &                   & 224$\times$224$\times$3 \\
                                \cline{2-4}
                                & Convolution    & 3$\times$3, stride=2 & 112$\times$112$\times$32 \\
                                \cline{2-4}
                                & Maxpooling     & 3$\times$3, stride=2 & 56$\times$56$\times$32  \\
\hline
\multirow{2}{*}{\textbf{Stage1}} & Downsample   & Bottleneck &  28$\times$28$\times$4c\\
                                \cline{2-4}
                                 & Feature enhancement      & Res2Blocks$\times$3 & 28$\times$28$\times$4c\\
\hline
\multirow{2}{*}{\textbf{Stage2}} & Downsample               & Bottleneck          &  14$\times$14$\times$8c\\
                                \cline{2-4}
                                 & Feature enhancement      & Res2Blocks$\times$7 & 14$\times$14$\times$8c\\
\hline
\multirow{2}{*}{\textbf{Stage3}} & Downsample               & Bottleneck          &  7$\times$7$\times$16c\\
                                \cline{2-4}
                                 & Feature enhancement      & Res2Blocks$\times$3 & 7$\times$7$\times$16c\\

\hline
\multirow{2}{*}{\textbf{Stage4}} & Average pooling    & 7$\times$7, stride=1          &  1$\times$1$\times$16c\\
                                \cline{2-4}
                                 & Convolution      & 1$\times$1, stride=1          & 1000-d\\
\bottomrule[2pt]
\end{tabular}}
\end{table}

High efficiency with strong feature representation ability is fundamental to lightweight accurate object detectors. As mentioned in the introduction, although many state-of-the-art algorithms borrow classification networks transferred directly from ImageNet pre-trained models as the backbone, there are obvious gaps between classification networks and the detection backbone. To avoid resulting in suboptimal architectures, we propose a new ResNet-like lightweight backbone network based on design principles for detection networks. Table~\ref{tab:Res2NetLite} lists in detail
the structure of our proposed backbone, \textit{Res2NetLite}.

\begin{figure}[t]
\begin{minipage}[t]{0.38\linewidth}
  \centering
  \centerline{\includegraphics[width=3.6cm]{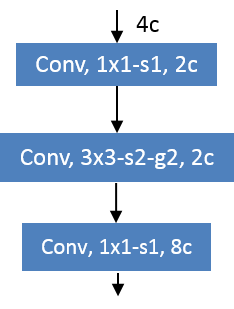}}
  \centerline{\small{(a) Bottleneck}}
\end{minipage}
\hfill
\begin{minipage}[t]{0.60\linewidth}
  \centering
  \centerline{\includegraphics[width=4.5cm]{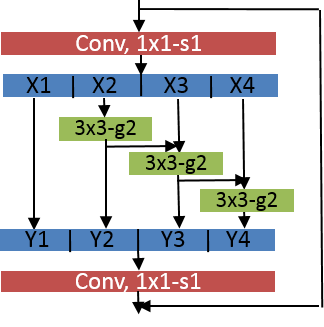}}
  \centerline{\small{(b) Res2Block}}
\end{minipage}
\caption{The detailed structures of Bottleneck and Res2Block used in Res2NetLite.}
\label{fig:BottleneckVSRes2Block}
\end{figure}

In ``Stage0'', Res2NetLite first quickly down-samples the input resolution to $1/4\times1/4$ and expands the feature dimension to $32$ through a simple batchnorm-convolution-maxpooling combination. Then, three stages are concatenated to form the main part of the network. In each stage, following one bottleneck module (Fig.~\ref{fig:BottleneckVSRes2Block}(a)), several repeated Res2Blocks~\cite{Res2Net} ($3$, $7$ and $3$ for Stage 1, 2 and 3, respectively) enhance the feature representation ability gradually. After each stage, the feature resolution is halved but the dimension is doubled. The $c$ is a feature dimension hyper-parameter to control the trade-off between network efficiency and accuracy. Finally, ``Stage4'' is just a conventional global pooling together with a $1\times1$ convolution layer.

\textbf{Suitable for detection.}
Different from classification networks, who are only concerned about the feature representation ability of the last convolution layer, pyramidal feature detection backbones care more about that of early stage features. So based on the basic consensus, we broaden the dimensions of low-level features deliberately as much as possible. For instance, in terms of Res2NetLite72 (c$=$72), the dimensions of the four pyramidal features used for anchor refinement in Fig.~\ref{fig:overall_framework} are $\{576,1152,512,512\}$ (the last convolution layers of ``Stage2'' and ``Stage3'', as well as two extra bottleneck modules as shown in Fig.~\ref{fig:BottleneckVSRes2Block}(a)). Another important issue is that detection backbones require large receptive fields to capture sufficient contextual information for effective localization and classification of large objects in the early stage.
Therefore, based on the above two considerations, the key point of our proposed backbone is the exploitation of Res2Block~\cite{Res2Net}. As depicted in Fig.~\ref{fig:BottleneckVSRes2Block}(b), Res2Block splits features into several groups and constructs hierarchical feature connection in a single residual block. Compared with the well-known ResBlock~\cite{ResNet}, Res2Block achieves multi-scale feature fusion and receptive field expansion but has less computational complexity.

\textbf{High efficiency.}
In~\cite{ShuffleNetV2}, Ma \etal derived several practical guidelines for an effective network architecture. In addition to the computational complexity of the network, two other factors---memory access cost (MAC) and degree of parallelism play the crucial role in affecting the exact running speed, especially on CPUs. Following the guidance, we strictly ensure the input and output channels are the same for Res2Blocks, which minimizes the MAC. On the other hand, we employ maximal $g=2$ (as shown in Fig.~\ref{fig:BottleneckVSRes2Block}) for all group convolutions in Bottlenecks and Res2Blocks since deep-wise convolutions are still not well optimized in CPU devices. Finally, we do not add any fragment structure, like inception or squeeze-and-excitation modules, in Res2NetLite. Generally speaking, our proposed backbone is a kind of straightforward and elegant network without any bypass branches. We enhance the feature representation ability simply by broadening feature dimensions and fusing multi-scale features inside each single block.
\subsection{Lightweight detection head and feature fusion}
\begin{figure}[t]
\begin{minipage}[t]{0.45\linewidth}
  \centering
  \centerline{\includegraphics[width=4.0cm]{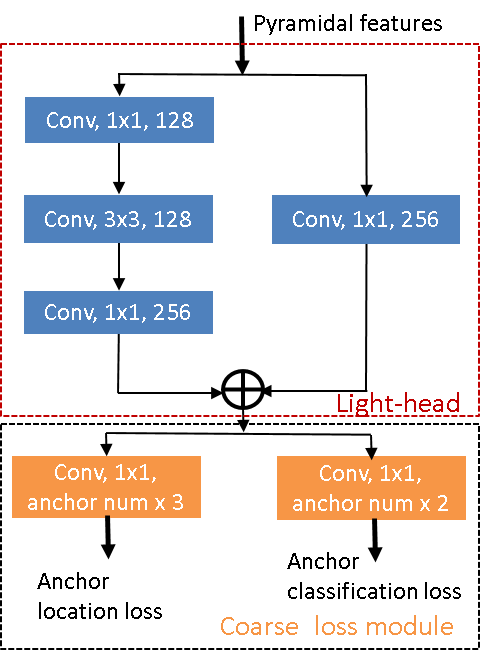}}
  \centerline{\small{(a) Light-head detection part}}
\end{minipage}
\hfill
\begin{minipage}[t]{0.51\linewidth}
  \centering
  \centerline{\includegraphics[width=4.4cm]{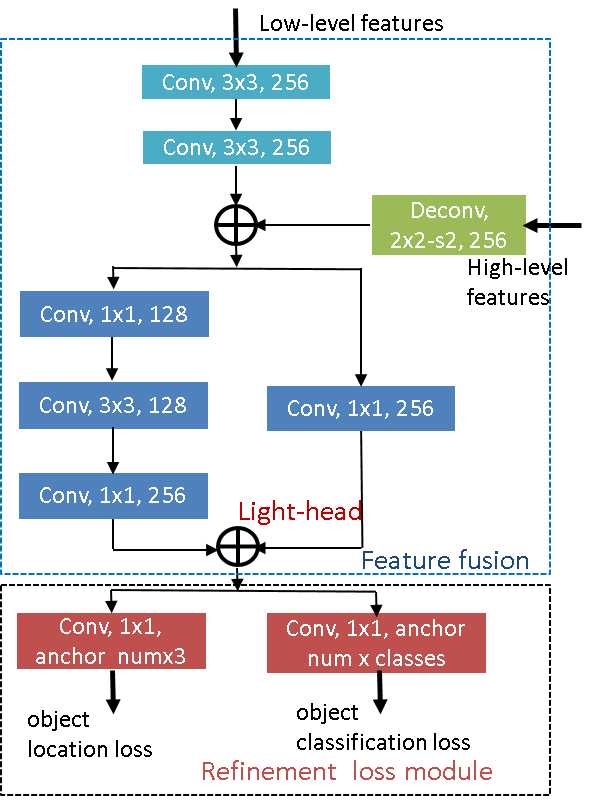}}
  \centerline{\small{(b) Light-head feature fusion}}
\end{minipage}
\caption{lightweight detection head for coarse losses and refined losses modules. The key component is the Light-head module.}
\label{fig:LightHead}
\end{figure}

To match the backbone capability, we propose the corresponding lightweight detection head and feature fusion module as illustrated in Fig.~\ref{fig:LightHead}. The key component is the residual-like \textit{Light-head} module as shown in Fig.~\ref{fig:LightHead}(a). Instead of original plain $3\times3$ convolution layers, the Light-head module further fuses high-level features with different receptive fields in a multi-scale two-pass way (a bottleneck structure and a $1\times1$ convolution layer). Then, this module also enables us to utilize only $1\times1$ convolutions to directly output location and class predictions.

\section{Training strategy improvements}
\label{sec:TrainingStrategy}
\subsection{Loss function}
The overall loss $L$ of RefineDetLite consists of four parts: location and classification losses for both the ARM (coarse losses) and ODM (refined losses), as shown in Eq.~\ref{eq:overall_loss}.
\begin{equation}\label{eq:overall_loss}
L = \lambda^{arm}_{loc}L_{loc}^{arm} + \lambda^{arm}_{cls}L_{cls}^{arm} + \lambda^{odm}_{loc}L_{loc}^{odm} + \lambda^{odm}_{cls}L_{cls}^{odm},
\end{equation}
where the $\lambda$s are the weighted coefficients and typically fixed to $1$, empirically. In most state-of-the-art papers, the location loss is approximated by the smooth L1 loss. But recently, Rezatofighi \etal proved that directly optimizing IOU metric is a better approach and accordingly proposed a GIOU loss~\cite{GIOU}. In terms of the classification loss, cross entropy loss is the most widely used method. In this paper, we propose to adopt a weighted KL-divergence loss as the classification loss for ODM as follows:
\begin{equation}\label{eq:loss_cls}
\begin{aligned}
 L_{cls}^{odm}\left(\mathbf{X},\mathbf{Y}\right) &= \frac{1}{N}\sum_{j=0}^{N-1}\eta_j\cdot L_{cls}(\mathbf{x}_j,\mathbf{y}_j) \\
                                            &= \frac{1}{N}\sum_{j=0}^{N-1}\eta_j\!\cdot\! \left(\sum_{i=0}^{K}w_{i}\!\cdot\! y_{ji}\!\cdot\! \!\log\! \frac{y_{ji}}{{\rm softmax}(x_{ji})}\right),
\end{aligned}
\end{equation}
where
$N$ is the mini-batch size, $j$ is the index of the anchor in the mini-batch, $\mathbf{x}_j$ is the network output predictions, $\mathbf{y}_j=(y_{j0},y_{j1},\cdots,y_{ji},\cdots,y_{jK})$ is the soft ground-truth label for anchor $j$, $K$ is the number of classes ($K-1$ foreground classes together with one background class), $w_i$ is the class weight for class $i$ and $\eta_j$ is the anchor sample weight.

\subsection{IOU-guided loss}
Apparently, if the ground-truth label $\mathbf{y}_j$ is the traditional one-hot vector, Eq.~\ref{eq:loss_cls} is exactly the weighted sum of binary cross entropy losses. So what we concerned with is to reform the soft ground-truth label $\mathbf{y}_j$ to improve the final network detection accuracy.

In almost all one-stage detectors, an anchor will be assigned a hard label with the probability $1$ if the maximal IOU between the anchor and a ground-truth box is above a threshold ($0.5$ for instance). But unfortunately, the detector cannot always optimize the candidate anchor perfectly, which means the final IOU can not be exactly equal to $1$ in most situations. So, this easily-neglected discrepancy may cause inconsistency when jointly optimizing localization and classification losses. Finally, it will lead to a phenomenon that some badly localized bounding boxes are classified as foreground objects with very high confidence.

To alleviate this inconsistency, we propose a so-called IOU-guided loss. Based on the above analysis, we are motivated to replace the one-hot vector $\mathbf{y}_j$ with soft labels. Denote the hard label for anchor $j$ as $t_j$ ($t_j=\{0,1,\cdots,K\}$ (assuming that $t_j=0$ means negative sample). The key item $y_{jt_j}$ in vector $\mathbf{y}_j$ is calculated as a function of ${\rm \widehat{IOU}}_j$ as
\begin{equation}\label{eq:soft_label}
y_{jt_j}=\left\{
                \begin{aligned}
                1-\alpha(1-{\rm \widehat{IOU}}_j) &, & t_j > 0\,&{\rm(positive\, sample)}\\
                1-\alpha{\rm \widehat{IOU}}_j &, & t_j = 0\,&{\rm(negative\, sample)}
                \end{aligned}
        \right.,
\end{equation}
where $\alpha$ is a hyper-parameter. After adding the predicted offsets onto the pre-defined anchor $j$, we can re-calculate a new IOU value (${\rm \widehat{IOU}}_j$) between the refined anchor $j$ and the pre-assigned ground-truth box. Notice that ${\rm \widehat{IOU}}_j \in [0, 1]$, based on the bounding box prediction accuracy. Therefore, for a positive sample, if the post-calculated ${\rm \widehat{IOU}}_j$ is small, the ground-truth label value $y_{jt_j}$ decreases accordingly.

To normalize the soft label $\mathbf{y}_j$, the other items except $y_{jt_j}$ are formulated as $y_{ji} = \frac{1-y_{jt_j}}{K-1}$.
We also weight the anchor samples according to ${\rm \widehat{IOU}}_j$ as
\begin{equation}\label{eq:anchor_weight}
\eta_j = \frac{1}{1+\beta({\rm \widehat{IOU}}_j - 1)\cdot \mathbf{1}(t_j > 0)},
\end{equation}
where $\beta$ is another hyper-parameter.
The physical meaning of Eq.~\ref{eq:anchor_weight} is that harder positive samples (lower post-calculated IOU value) will be assigned higher weights.

\subsection{Dataset-aware classes-weighted loss}
In the past few years, many researchers focused on solving the imbalance problem between positive and negative samples for one-stage detectors~\cite{FocalLoss,GHMLoss}, but few works dealt with the imbalance problem among different classes in training data. Consequently, we propose a classes-weighted loss based on the statistics of the training dataset. Assuming the number of labeled boxes for class $i$ is $M_i$, the maximal and minimal number are $M_{max}$ and $M_{min}$, respectively. Therefore, the loss weight for class $i$ is computed as
\begin{equation}\label{eq:classes_weight}
w_i = \left\{
            \begin{aligned}
            \frac{W-1}{r_{max}^\gamma - 1}\cdot (r_i^\gamma - 1) + 1 &, & i > 0\, &{\rm (forground)}\\
            1 &, & i = 0\,&{\rm (background)}
            \end{aligned}
    \right.,
\end{equation}
where
$r_i=\frac{M_{max}}{M_i}$, $r_{max}=\frac{M_{max}}{M_{min}}$, and $\gamma$ and $W$ are two hyper-parameters. Obviously, the loss weight for the most frequent class is $1$ while the most rare class is $W$.

\subsection{Balanced multi-task learning}
As derived in Eq.~\ref{eq:overall_loss}, the overall loss is actually a weighted sum of different parts, but how to determine the optimal weighted coefficients is a challenge.

In~\cite{MultiTask}, Kendall \etal proposed to form a multi-task learning paradigm by considering homoscedastic uncertainty to automatically train loss weights for different computer vision tasks. Inspired by this idea, we can also view the object detection training process as a multi-task learning problem---regression and classification tasks. The advanced version of final loss function\footnote{For detailed derivation, please refer to~\cite{MultiTask}.} is approximated as
\begin{equation}\label{eq:multitask}
\begin{aligned}
L       &= \frac{1}{2\sigma_{1}^2}L_{loc}^{arm} + \frac{\log\sigma_{1}^2}{2} + \frac{1}{\sigma_{2}^2}L_{cls}^{arm} + \frac{\log\sigma_{2}^2}{2} \\
        &+ \frac{1}{2\sigma_{3}^2}L_{loc}^{odm} + \frac{\log\sigma_{3}^2}{2} + \frac{1}{\sigma_{4}^2}L_{cls}^{odm} + \frac{\log\sigma_{4}^2}{2},
\end{aligned}
\end{equation}
where $\{\sigma_{1}^2, \sigma_{2}^2, \sigma_{3}^2,  \sigma_{4}^2\}$ are the trainable uncertainty parameters. In practice, for the steady of overall loss during training, the trainable parameters consist of network parameters $\bm{\phi}$ and \textit{logarithmic form} of the four uncertainty parameters.

\section{Experiments}
\begin{table*}[t]
\centering
\caption{Detection result comparisons on MSCOCO online test-dev server, where time@CPU1 means running time tested by us based on open-source codes (All SSD and RefineDet-based networks~\cite{urlSSD}, Tiny-DSOS~\cite{urlTinyDSOD}, ThunderNet~\cite{urlThunderNet}, Pelee~\cite{urlPelee}, YOLOV3~\cite{urlYOLOV3}) on Intel i7-6700@3.40GHz and time@CPU2 means running time claimed by the original authors on different platforms. Bold fonts indicate our proposed algorithms. \textbf{RefineDetLite++} means the model trained on the\textit{ advanced configuration}.}\label{tab:Results}
\resizebox{1.99\columnwidth}{!}{
\begin{tabular}{c|c|c|c|c|c|c|c|c|c|c}
\hline
Algorithms                      & input size     & Backbone           & AP & AP$_{50}$ & AP$_{75}$ & AP$_S$ & AP$_M$ & AP$_L$ & time@CPU1 & time@CPU2\\
\hline
\textit{one-stage lightweight:}             &                &                    &    &         &         &        &       &       &  &\\
SSD~\cite{SSD}                  &300$\times$300  &MobileNet           & 19.3 &  -  &  -    &  -   &  -  &   - & 128ms & - \\
SSDLite~\cite{MobileNetV2}      &320$\times$320  &MobileNet           & 22.2 &  -  &  -    & -    &  -  &   - & 125ms & 270ms (Pixel 1)\\
SSDLite~\cite{MobileNetV2}      &320$\times$320  &MobileNetV2         & 22.1 &  -  &  -    & -    &  -  &   -  & 120ms &200ms (Pixel 1)\\
Pelee~\cite{PeleeNet}           &304$\times$304  &PeleeNet            & 22.4   &  38.3  & 22.9     &  -    & -      &  -   &140ms & 149ms (intel i7)\\
Tiny-DSOD~\cite{Tiny-DSOD}      &300$\times$300  &DDB-Net+D-FPN       & 23.2   &  40.4  &  22.8    &  -    &   -    &  -    &180ms  &-\\
\hline
\hline
\textit{RefineDet-based lightweight:}  &                &                    &    &         &         &        &        &         &  &\\
RefineDet~\cite{RefineDet}      &320$\times$320  &MobileNet           & 24.3   &  43.0  &  24.4    & 7.6   & 26.5   & 38.7  & 168ms & - \\
RefineDet~\cite{RefineDet}      &320$\times$320  &MobileNetV2         & 24.8   &  42.9  &  25.4    & 7.6   & 26.5   & 40.4    & 163ms &- \\
RefineDet~\cite{RefineDet}      &320$\times$320  &MobileNetV3         & 22.1   &  40.1  &  23.3    & 6.5   & 23.6   & 35.7   & 150ms &- \\
RefineDet~\cite{RefineDet}      &320$\times$320  &ShuffleNetV2        & 21.1   &  38.3  &  21.5    & 5.8   & 22.8   & 34.3    & 129ms &- \\
\hline
\hline
\textit{two-stage lightweight:}             &               &                    &    &         &         &        &        &         &  &\\
FPNLite (@ 64)~\cite{NAS-FPN}       &320$\times$320 & MobileNetV2        & 22.7   &   -     &   -   &    -  &   -  &   -    & - &  192ms (Pixel 1)\\
FPNLite (@ 128)~\cite{NAS-FPN}      &320$\times$320 & MobileNetV2        & 24.3   &   -     &    -  &    -  &   -  &   -       & - & 264ms (Pixel 1)\\
NAS-FPNLite (3 @ 48)~\cite{NAS-FPN} &320$\times$320 & MobileNetV2        & 24.2   &   -     &  -    &   -   &   -  &  -        & - & 210ms (Pixel 1)\\
NAS-FPNLite (7 @ 64)~\cite{NAS-FPN} &320$\times$320 & MobileNetV2        & 25.7   &   -     &   -   &   -   &   -  &   -        & - & 285ms (Pixel 1)\\
ThunderNet~\cite{ThunderNet}        &320$\times$320 & SNet535            & 28.0   &   46.2  &  29.5 &  -    &   -  &   -       & 146ms &172ms (Snapdragon 845)\\
\hline
\hline
\textit{one-stage classical:}              &               &                    &    &         &         &        &        &        &  &\\
SSD~\cite{SSD}                  &300$\times$300  &VGG                 &  25.1  &    43.1     & 25.8        &  6.6       & 25.9   &  41.4  & 1250ms  &-\\
SSD~\cite{SSD}                  &321$\times$321  &ResNet101           &  28.0  &    45.4     & 29.3        &  6.2       & 28.3   &  49.3  & 1000ms  &-\\
YOLOV3~\cite{YOLOV3}            &320$\times$320  &DarkNet53           &  28.2  &    51.5    & -  & -  &-  &   -    & 1300ms &- \\
\hline
\hline
\textbf{RefineDetLite}           &\textbf{320$\times$320} & \textbf{Res2NetLite72}      & \textbf{26.8}   &   \textbf{46.6}  & \textbf{27.4} &  \textbf{7.4}   &  \textbf{27.7}  &  \textbf{42.4}   &\textbf{130ms} & -\\
\textbf{RefineDetLite++}  &\textbf{320$\times$320} & \textbf{Res2NetLite72}      & \textbf{29.6}   &   \textbf{47.4}  &  \textbf{31.0} & \textbf{9.1}   &  \textbf{30.8}  &  \textbf{45.5}  & \textbf{131ms} &-\\
\hline
\end{tabular}}
\end{table*}
In this section, we elaborately present the experimental results and ablation studies on the MSCOCO~\cite{MSCOCO} dataset to demonstrate the effectiveness of our proposed RefineDetLite and training strategies.
\subsection{Training and inference details}

\textbf{Basic configuration.}
To demonstrate the effectiveness of our network architecture, we set a basic configuration for fair comparisons with other state-of-the-art networks.
All experiments are conducted on 4 Nvidia P40 GPUs with a batchsize of $128$. We choose SGD with a weight decay of $0.0005$ and momentum of $0.9$ as our optimizer. We set the learning rate to $4\times10^{-3}$ for the first $150$ epochs, and decay it to $4\times10^{-4}$ and $4\times10^{-5}$ for training another $50$ and $50$ epochs, respectively. RefineDetLite72 is adopted as our backbone. For data augmentation, we strictly follow the instruction of SSD~\cite{SSD}. The loss functions for regression and classification are smooth L1 loss and cross-entropy loss, respectively. The weighted coefficients in Eq.~\ref{eq:overall_loss} are all fixed to $1$. We choose OHEM with a conventional positive-negative ratio of $1:3$ as our training strategy. During inference, we post-process the network outputs by normal nms.

\textbf{Advanced configuration.}
The advanced configuration is set to evaluate the training strategies proposed in Sec.~\ref{sec:TrainingStrategy} and improve the detection accuracy without apparent extra time-consumption as far as possible. GIOU loss~\cite{GIOU} is chosen as the regression loss function. On the other hand, we adopt the classification loss functions and multi-task training strategies proposed in Sec.~\ref{sec:TrainingStrategy}. Specifically, the hyper-parameters $\alpha, \beta, \gamma$ and $W$ are set to $0.25, 0.90, 3/4$ and $10$ experimentally. Correspondingly, the learning rate is adjusted as $\{10^{-2}, 10^{-3}, 10^{-4}\}$ for the same epochs described in the basic configuration. Finally, soft-nms~\cite{soft-nms} is employed as the post-processing approach. The data augmentation and optimizer are kept the same as with the basic configuration. Controlled experiments are presented in Sec.~\ref{sec:AblationStudy} to evaluate the effectiveness of each component.

\subsection{Results on MSCOCO}
MSCOCO~\cite{MSCOCO} is the most widely-used object detection evaluation dataset, which consists of 118,287 training samples (\verb"trainval35k"), 5,000 validation samples (\verb"val5k") and 40,670 test samples (\verb"test-dev") in its 2017 version. We trained all networks on \verb"trainval35k" and evaluated on \verb"test-dev" for fair comparisons.

All models are trained on Pytorch~\cite{Pytorch} and inferred on Caffe2 converted by ONNX¡¤\cite{ONNX}. As discussed in the introduction, since network FLOPs are not proportional to the exact speed due to many factors, the network efficiency measured by FLOPs is not an accurate criterion. What people are really concerned with is the actual network running speed on CPU devices. Therefore, we re-implement some state-of-the-art algorithms and test their time efficiency (ms/pic) by averaging the total running time of $100$ images on a single-thread Intel i7-6700@3.40GHz GPU for fair comparisons. Additionally, we also show the speed claimed by the authors tested on different platforms as references. The detailed comparisons are listed in Table~\ref{tab:Results}.

Without any bells and whistles, our proposed RefineDetLite trained on the basic configuration achieves $26.8$ mAP at a speed of $130$ ms/pic, which surpasses state-of-the-art \textit{one-shot lightweight} detectors (SSD~\cite{SSD}, SSDLite~\cite{MobileNetV2}, Pelee~\cite{PeleeNet} and Tiny-DSOD~\cite{Tiny-DSOD}). Specifically, RefineDetLite achieves SSDLite-MobileNet level speed with a significant 4.6 mAP improvement. In addition, it outperforms Pelee and Tiny-DSOD in both accuracy and speed.

We also conducted comparison experiments on some state-of-the-art \textit{two-stage lightweight} detection algorithms. The most competitive approach is the elaborately designed ThunderNet-SNet535~\cite{ThunderNet}, which achieves $28.0$ mAP (which surpasses our basic version $26.8$, but is less than the advanced version $29.6$). Since no official code was released by the authors, we did the speed comparison experiments based on a third-party re-implementation~\cite{urlThunderNet}. Besides this, NAS-FPNLites and FPNLites~\cite{NAS-FPN} are lightweight version of two-stage detectors based on NAS-FPN and FPN~\cite{FPN}, respectively, but the accuracies are still lower than RefineDetLite. Since no open-released code can be referenced to do speed comparisons, we can only perform indirectly reasoning according to the speed of SSDLites and NAS-FPNLites both on a Pixel 1 CPU.

For further study of the effectiveness of our proposed backbone and light-head structures, we implemented RefineDet with other lightweight backbones (MobileNets serials~\cite{MobileNet,MobileNetV2,MobileNetV3} and ShuffleNetV2~\cite{ShuffleNetV2}). The statistics in Table~\ref{tab:Results} reveal that RefineDetLite surpasses all MobileNet-based RefineDet in both accuracy and efficiency. Additionally, at the same running speed level, it beats RefineDet-ShuffleNetV2 by a huge $5.7$ mAP gain.

Finally, by adding training and inference strategies (soft-nms, GIOU and the three new strategies proposed in Sec.~\ref{sec:TrainingStrategy}), we can further improve the RefineDetLite detection accuracy from $26.8$ to $29.6$ with little extra time consumption. One issue that deserves to be mentioned is that RefineDetLite can achieve similar detection accuracy level of some classical one-stage detectors (SSD-VGG, SSD-ResNet101 and YOLOV3) with an $8$--$10$ times speed increase.
\begin{figure*}[t]
\begin{minipage}[t]{0.195\linewidth}
  \centering
  \centerline{\includegraphics[width=3.5cm]{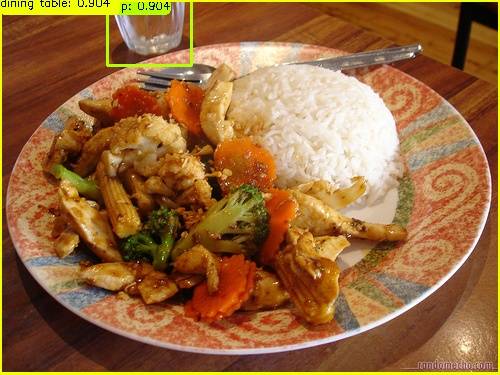}}
\end{minipage}
\hfill
\begin{minipage}[t]{0.195\linewidth}
  \centering
  \centerline{\includegraphics[width=3.5cm]{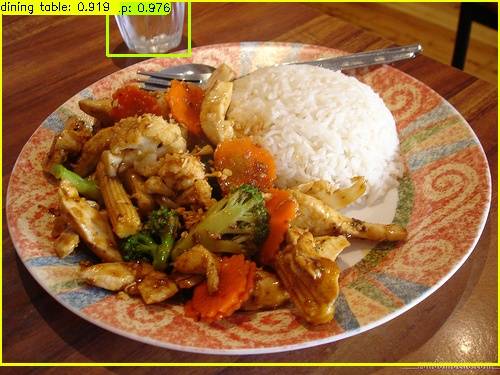}}
\end{minipage}
\hfill
\begin{minipage}[t]{0.195\linewidth}
  \centering
  \centerline{\includegraphics[width=3.5cm]{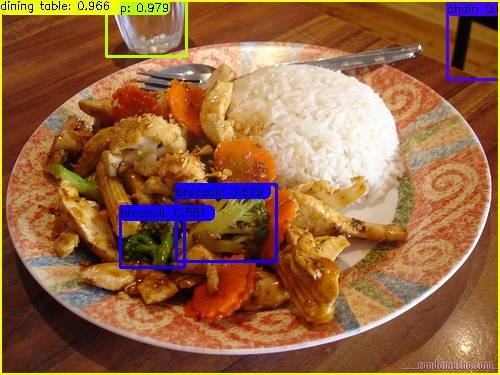}}
\end{minipage}
\hfill
\begin{minipage}[t]{0.195\linewidth}
  \centering
  \centerline{\includegraphics[width=3.5cm]{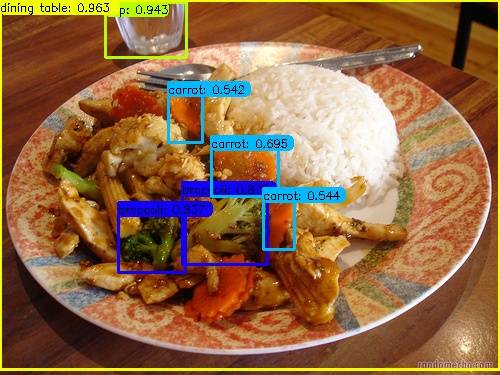}}
\end{minipage}
\hfill
\begin{minipage}[t]{0.195\linewidth}
  \centering
  \centerline{\includegraphics[width=3.5cm]{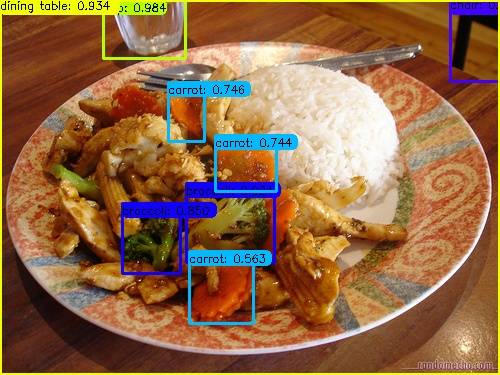}}
\end{minipage}\\
\begin{minipage}[t]{0.195\linewidth}
  \centering
  \centerline{\includegraphics[width=3.5cm]{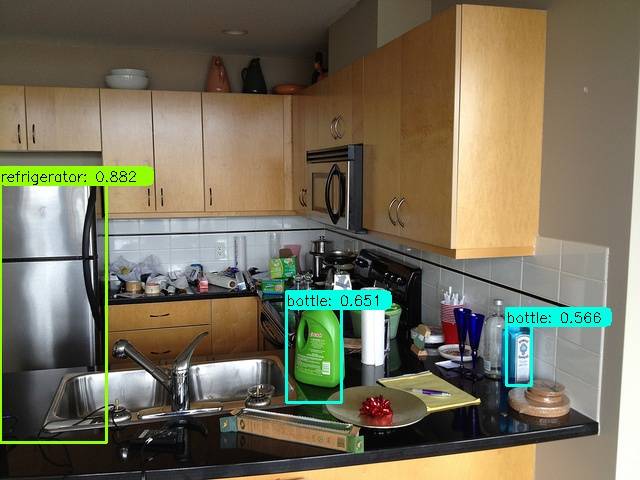}}
\end{minipage}
\hfill
\begin{minipage}[t]{0.195\linewidth}
  \centering
  \centerline{\includegraphics[width=3.5cm]{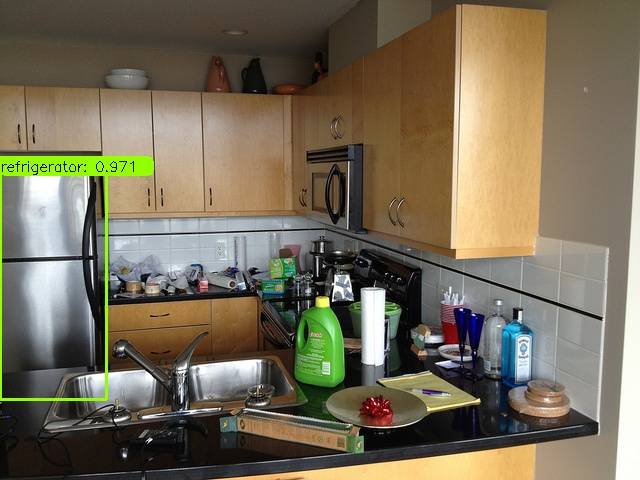}}
\end{minipage}
\hfill
\begin{minipage}[t]{0.195\linewidth}
  \centering
  \centerline{\includegraphics[width=3.5cm]{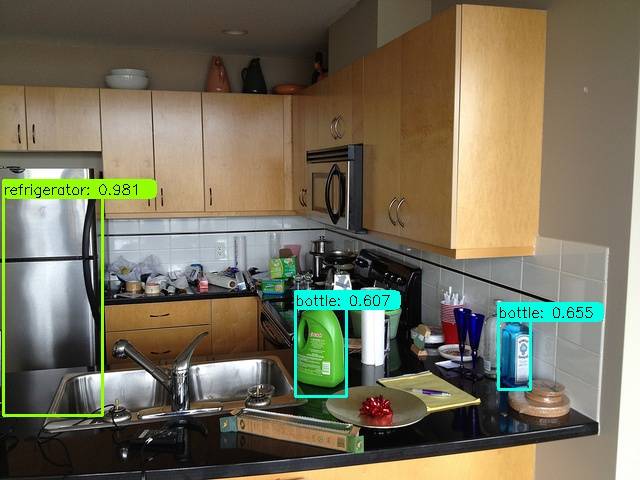}}
\end{minipage}
\hfill
\begin{minipage}[t]{0.195\linewidth}
  \centering
  \centerline{\includegraphics[width=3.5cm]{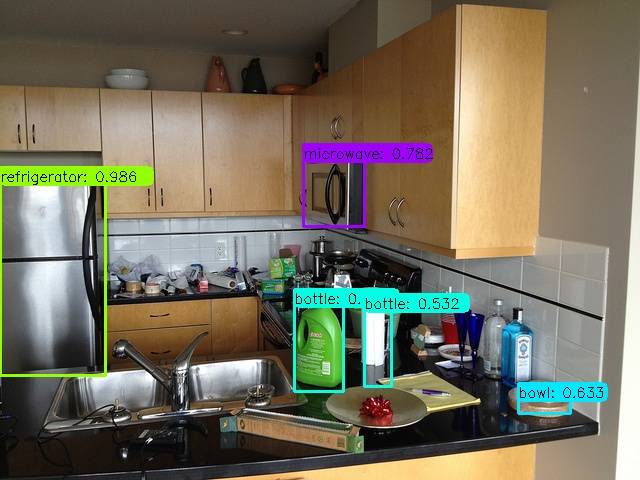}}
\end{minipage}
\hfill
\begin{minipage}[t]{0.195\linewidth}
  \centering
  \centerline{\includegraphics[width=3.5cm]{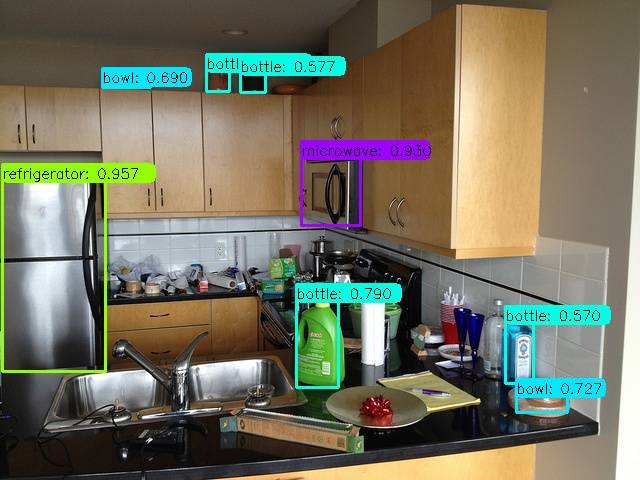}}
\end{minipage}\\
\begin{minipage}[t]{0.195\linewidth}
  \centering
  \centerline{\includegraphics[width=3.5cm]{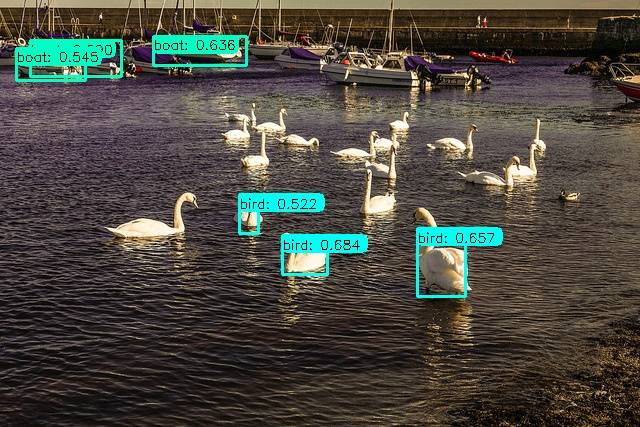}}
  \centerline{\small{(a) SSDLite+MobileNet}}
\end{minipage}
\hfill
\begin{minipage}[t]{0.195\linewidth}
  \centering
  \centerline{\includegraphics[width=3.5cm]{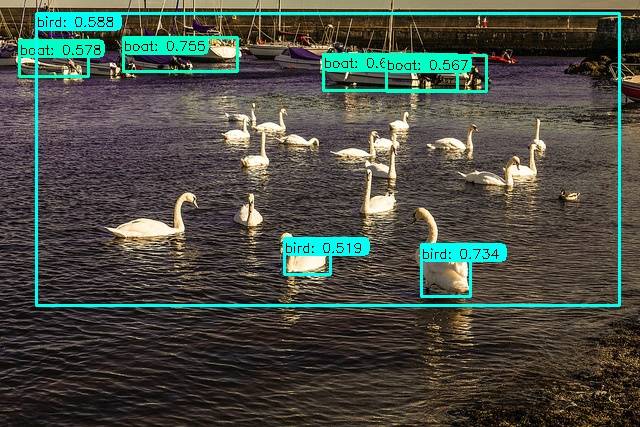}}
  \centerline{\small{(b) RefineDet+ShuffleNetV2}}
\end{minipage}
\hfill
\begin{minipage}[t]{0.195\linewidth}
  \centering
  \centerline{\includegraphics[width=3.5cm]{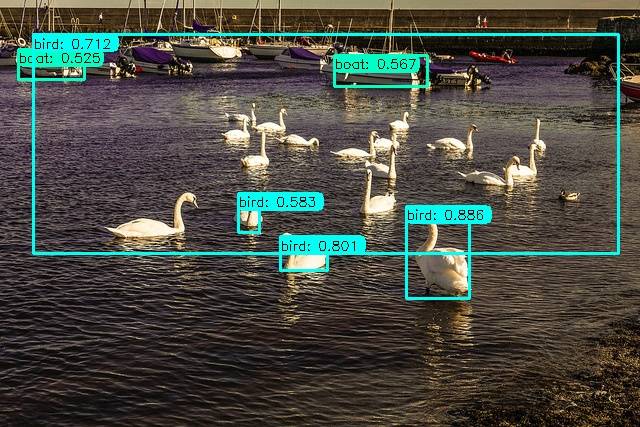}}
  \centerline{\small{(c) SSD+VGG}}
\end{minipage}
\hfill
\begin{minipage}[t]{0.195\linewidth}
  \centering
  \centerline{\includegraphics[width=3.5cm]{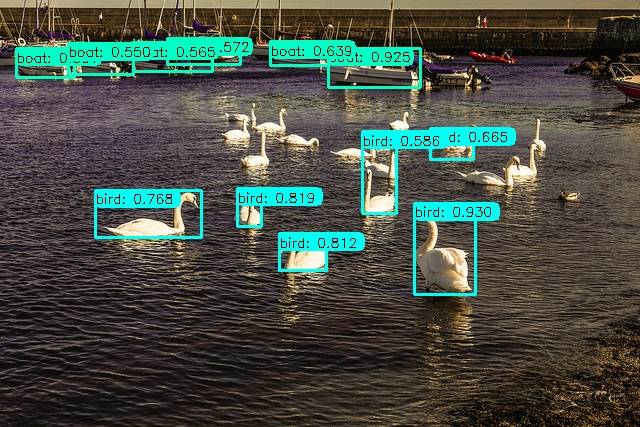}}
  \centerline{\small{(d) RefineDetLite}}
\end{minipage}
\hfill
\begin{minipage}[t]{0.195\linewidth}
  \centering
  \centerline{\includegraphics[width=3.5cm]{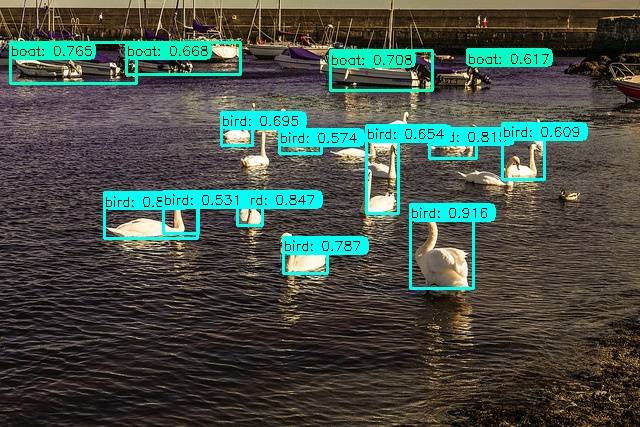}}
  \centerline{(e) RefineDetLite++}
\end{minipage}
\caption{Visualization comparisons among different models.}
\label{fig:visualization}
\end{figure*}

Additionally, we visualize the comparison results among different models---SSDLite+MobileNet, RefineDet+ShuffleNetV2, SSD+VGG, RefineDetLite and RefineDetLite++ as in Fig.~\ref{fig:visualization}. The visualization results clearly show that the RefineDetLite can detect much more small objects (\eg, carrots, bottles and birds) compared with SSDLite+MobileNet and RefineDet+ShuffleNetV2 at almost the same running speed ($130$ ms/pic). It also outperforms SSD+VGG with a considerable $10$ times of speed increase. Furthermore, by adding the proposed training strategies, the detection accuracy and recall of RefineDetLite++ surpasses RefineDetLite significantly.
\subsection{Ablation study}
\label{sec:AblationStudy}
\begin{table*}[t]
\centering
\caption{Ablation studies on proposed detection modules and training strategies. Bold fonts indicate our proposed modules.}\label{tab:ablation_study}
\resizebox{1.99\columnwidth}{!}{
\begin{tabular}{c|c|c|c|c|c|c|c|c|c|c|c}
\hline
models            & \textbf{Res2Block} & \textbf{Light-head} & soft-nms & GIOU & \textbf{classes weights} & \textbf{IOU guided} &\textbf{ Multi-task} & AP & AP$_{50}$ & AP$_{75}$ & time@CPU1 \\
\hline
(a) Baseline                     &                &                 &                 &              &          &            &             &  24.7  &   43.0     &   25.1     &   150ms   \\
(b)                              & $\checkmark$   &                 &                 &              &          &            &             &  25.7  &   43.9     &   26.7     &  140ms      \\
(c) \textbf{RefineDetLite}        & $\checkmark$   &  $\checkmark$   &                 &              &          &            &             &  \textbf{26.8}   &   \textbf{46.6}  & \textbf{27.4}& \textbf{130ms}               \\
\hline
(d)                              & $\checkmark$   &  $\checkmark$   & $\checkmark$    &              &          &            &             &  27.2   & 46.0 &  28.4   & 131ms     \\
(e)                              & $\checkmark$   &  $\checkmark$   & $\checkmark$    &              & $\checkmark$  &     &               &  27.4   & 46.2 &  28.8   & 131ms      \\
(f)                              & $\checkmark$   &  $\checkmark$   & $\checkmark$    &              &          &  $\checkmark$ &          &  27.5   & 46.2 &  28.9     &  131ms  \\
(g)                              & $\checkmark$   &  $\checkmark$   & $\checkmark$    &              &          &            & $\checkmark$ & 27.7   & 46.5 &  29.0     &  131ms     \\
\hline
(h)                              & $\checkmark$   &  $\checkmark$   & $\checkmark$    & $\checkmark$ &          &            &             &  28.2   & 46.4 &  29.4   & 131ms         \\
(i)                              & $\checkmark$   &  $\checkmark$   & $\checkmark$    & $\checkmark$ &$\checkmark$   &            &             & 28.5 &  46.8    & 29.8    &  131ms \\
(j)                              & $\checkmark$   &  $\checkmark$   & $\checkmark$    & $\checkmark$ &          &  $\checkmark$   &             & 28.6 &  46.8     &  29.9    &  131ms  \\
(k)                              & $\checkmark$   &  $\checkmark$   & $\checkmark$    & $\checkmark$ &          &            & $\checkmark$     & 28.7 &  47.0     &   30.2     &  131ms   \\
\hline
(l)                              & $\checkmark$   &  $\checkmark$   & $\checkmark$    & $\checkmark$ &$\checkmark$   &  $\checkmark$   &       & 29.0   &   47.3    &   30.1   & 131ms   \\
(m)                              & $\checkmark$   &  $\checkmark$   & $\checkmark$    & $\checkmark$ &          &  $\checkmark$   & $\checkmark$     & 29.3&   47.4    & 30.5   &131ms \\
(n)                              & $\checkmark$   &  $\checkmark$   & $\checkmark$    & $\checkmark$ & $\checkmark$  &            & $\checkmark$     & 29.2&   47.4    &  30.4      &131ms \\
(o) \textbf{RefineDetLite++} & $\checkmark$   &  $\checkmark$   & $\checkmark$    & $\checkmark$ & $\checkmark$  &  $\checkmark$   & $\checkmark$     &  \textbf{29.6}   &   \textbf{47.4}  &  \textbf{31.0}  &\textbf{131ms}            \\
\hline
\end{tabular}}
\end{table*}
\textbf{Network architectures.} First, we evaluate the effectiveness the two key components (\textit{Res2Block} and \textit{Light-head}) of our proposed RefineDetLite architectures. We set up a baseline architecture, just simply replacing all Res2Blocks in the backbone Res2NetLite by the classical ResBlocks~\cite{ResNet} and replacing all Light-head modules in the detection part and feature fusion by simple $3\times3$ convolutions. Experimental results in rows (a)--(c) of Table~\ref{tab:ablation_study} show that the two components achieve $1.0$ and $1.1$ mAP improvements, separately. Simultaneously, they save the running time $10$ms and $10$ms, separately. Therefore, we can draw the conclusion that Res2Block and Light-head make great contributions to improving both the accuracy and efficiency of the detector.

\textbf{Training and inference strategies.}
Based on the RefineDetLite architecture trained on the basic configuration, we further conducted more controlled experiments to evaluate the effectiveness of each training and inference strategy, including the three strategies proposed in Sec.~\ref{sec:TrainingStrategy}---classes-weighted loss, IOU-guided loss and multi-task training and two orthogonal approaches proposed by other researchers: soft-nms and GIOU loss.

Rows (d) and (h) in Table~\ref{tab:ablation_study} demonstrate the consistent improvement of the well-known soft-nms and GIOU loss, which will be exploited as standard components for future experiments. Rows (e)--(g) reveal that the three proposed strategies---classes-weighted loss, IOU guided loss and multi-task training achieve $0.2$, $0.3$ and $0.5$ mAP gains, respectively,  after using soft-nms. Furthermore, after employing GIOU loss, the three components still achieve $0.3$, $0.4$ and $0.5$ mAP improvements (rows (i)--(k)).

Statistics in rows (l)--(o) demonstrate the orthogonality of the three proposed strategies. Experiments show steady accuracy increasing after combinations of any components. Finally, when integrating all strategies together, we can achieve a $29.6$ mAP, which is a state-of-the-art detection accuracy at the 150 ms/pic running speed level.

\section{Conclusion}

In this paper, we proposed a lightweight efficient and accurate one-shot object detection framework. After analyzing the gaps between classification networks and detection backbones, and following the design principles of efficient networks, we designed a residual-like lightweight backbone---Res2NetLite, enlarging the receptive fields and feature dimensions of early stages, together with a corresponding light-head detection parts. In addition, we investigate the weaknesses of current one-stage training process, hence develop three improved training strategies: IOU-guided loss, classes-weighted loss and balanced multi-task training method. Without any bells and whistles, our proposed RefineDetLite achieves $26.8$ mAP on MSCOCO \verb"test-dev" at a speed of 130 ms/pic. By adding the orthogonal training and inference strategies, the advanced version---RefineDetLite++ can further achieve $29.6$ mAP with little extra time consumption.

{\small
\bibliographystyle{ieee_fullname}
\bibliography{egbib}
}

\appendixpage
\appendix
\section{Example of classes-weighted loss}
\begin{figure}[ht]
  \centering
  \centerline{\includegraphics[width=8.5cm]{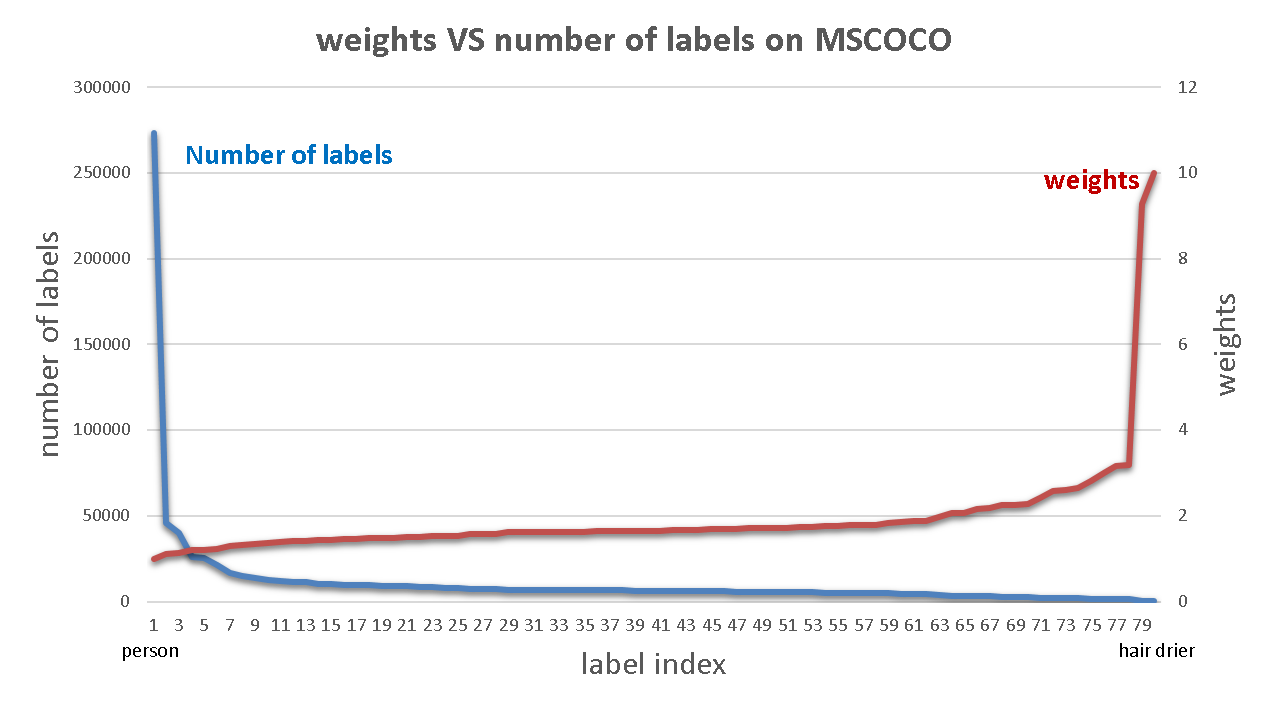}}
\caption{Number of labels VS weights on MSCOCO trainval35k dataset.}
\label{fig:MSCOCOexample}
\end{figure}

Here we show the example of classes-aware weights for MSCOCO~\cite{MSCOCO} \verb"trainval35k" dataset. Fig.~\ref{fig:MSCOCOexample} illustrates the curves of weights VS number of labeled boxes for all 80 COCO classes. Specifically, for the most frequent class---\textit{person} which contains $273468$ labeled boxes, the weight is $1$. While for the most rare class---\textit{hair drier} which contains only $209$ labeled boxes, the weight is $10$. We also test other hyper-parameters ($\gamma$ and $W$) in Eq.~\ref{eq:classes_weight} of original paper, but finally we chose $\gamma=3/4$ and $W=10$ experimentally based on the MSCOCO dataset.

\section{Derivation of balanced multi-task training}
Without loss of generality, we derive only the object detection loss $L^{odm}$ and omit the sample index subscript $j$. Following the guidance of~\cite{MultiTask}, minimizing the loss equals to minimizing the negative log likelihood as Eq.~\ref{eq:multitask_detail}.
\begin{figure*}[ht]
\begin{equation}\label{eq:multitask_detail}
\begin{aligned}
L^{odm} &= -\log P\left(\mathbf{g}, \mathbf{y} | \mathbf{f}^{\phi}(\mathbf{z})\right) \\
        &= -\log\left[P\left(\mathbf{g} | \mathbf{f}^{\phi}(\mathbf{z})\right)\cdot P\left(\mathbf{y} | \mathbf{f}^{\phi}(\mathbf{z})\right)\right]\quad(\rm{suppose~the~independence~of~ location~and~classification})\\
        &= -\log N(\mathbf{g}; \mathbf{f}^{\phi}(\mathbf{z}), \sigma_{loc}^2) + D_{KL}\left(\mathbf{y} || \frac{\mathbf{f}^{\phi}(\mathbf{z})}{\sigma_{cls}^2}\right)\quad(\rm{probability~ distributions~with~variances~\sigma_3^2~and~\sigma_4^2})\\
        &= \frac{1}{2\sigma_{3}^2}\left\|\mathbf{g} - \mathbf{f}^{\phi}(\mathbf{z})\right\| + \log\sigma_{3} + \frac{1}{2}\log2\pi + \sum_i y_i\log y_i -\sum_i y_i\log {\rm softmax}\left(\frac{\mathbf{f}^{\phi}_i(\mathbf{z})}{\sigma_{4}^2}\right) \\
        &\propto \frac{1}{2\sigma_{3}^2}\left\|\mathbf{g} - \mathbf{f}^{\phi}(\mathbf{z})\right\| + \log\sigma_{3} -\sum_i y_i\log {\rm softmax}\left(\frac{\mathbf{f}^{\phi}_i(\mathbf{z})}{\sigma_{4}^2}\right)\quad(\frac{1}{2}\log2\pi +\sum_i y_i\log y_i\rm{~is~nontrainable})\\
        &\approx \frac{1}{2\sigma_{3}^2}L_{loc}^{odm} + \frac{\log\sigma_{3}^2}{2} - \sum_i y_i\log {\rm softmax}\left(\frac{\mathbf{f}^{\phi}_i(\mathbf{z})}{\sigma_{4}^2}\right)\quad(L_{loc}^{odm}~\rm{is~approximated~as~L2~norm~loss})\\
        &\approx \frac{1}{2\sigma_{3}^2}L_{loc}^{odm} + \frac{\log\sigma_{3}^2}{2} + \frac{1}{\sigma_{4}^2}L_{cls}^{odm} + \frac{\log\sigma_{4}^2}{2}.
\end{aligned}
\end{equation}
\end{figure*}

\begin{figure*}[ht]
\begin{equation}\label{eq:approx}
\begin{aligned}
-\sum_i y_i\log {\rm softmax}\left(\frac{\mathbf{f}^{\phi}_i(\mathbf{z})}{\sigma_{4}^2}\right) &= -\sum_i y_i\log\frac{\exp\left(\frac{\mathbf{f}^{\phi}_i(\mathbf{z})}{\sigma_4^2}\right)}{\sum_c\exp\left(\frac{\mathbf{f}^{\phi}_c(\mathbf{z)}}{\sigma_4^2}\right)} \\
& =  -\frac{1}{\sigma_4^2}\sum_i y_i\log\left[\frac{\exp\left(\mathbf{f}^{\phi}_i(\mathbf{z})\right)}{\sum_c\exp\left(\mathbf{f}^{\phi}_c(\mathbf{z)}\right)}\cdot
\frac{\exp\left(\mathbf{f}^{\phi}_c(\mathbf{z})\right)}{\sum_c\exp\left(\frac{\mathbf{f}^{\phi}_c(\mathbf{z)}}{\sigma_4^2}\right)}\right]\\
& = -\frac{1}{\sigma_4^2}\sum_i y_i\left[\log\frac{\exp\left(\mathbf{f}^{\phi}_i(\mathbf{z})\right)}{\sum_c\exp\left(\mathbf{f}^{\phi}_c(\mathbf{z)}\right)}\right]
 + \sum_i y_i \log\left[ \frac{\exp\left(\mathbf{f}^{\phi}_c(\mathbf{z})\right)}{\left(\sum_c\exp\left(\frac{\mathbf{f}^{\phi}_c(\mathbf{z)}}{\sigma_4^2}\right)\right)^\frac{1}{\sigma_4^2}} \right]\\
 & \approx \frac{1}{\sigma_4^2}\left[\sum_i y_i \log\rm{softmax}\left(\mathbf{f}_i^{\bm{\phi}}(\mathbf{z})\right)\right]+ \log\left[ \frac{\exp\left(\mathbf{f}^{\phi}_c(\mathbf{z})\right)}{\left(\sum_c\exp\left(\frac{\mathbf{f}^{\phi}_c(\mathbf{z)}}{\sigma_4^2}\right)\right)^\frac{1}{\sigma_4^2}} \right]\\
 & \approx \frac{1}{\sigma_4^2}L_{cls}^{odm} + \log\sigma_4\quad(L_{cls}^{odm}\rm{~is~approximated~as~cross~entropy~loss~and~refer~to~\cite{MultiTask}})
\end{aligned}
\end{equation}
\end{figure*}

 In Eq.~\ref{eq:multitask_detail}, $\mathbf{g},\mathbf{y}$ are the ground-truth bounding boxes and class labels, respectively, $\mathbf{z}$ is the input image, $\mathbf{f}^{\phi}$ is the network with parameters $\phi$ and $\{\sigma_{3}, \sigma_{4}\}$ are the trainable uncertainty parameters. Based on the theorem of~\cite{MultiTask}, we assume that the regression and classification are two independent optimization processes and the final outputs are two probability distributions with random variables $\sigma_{3}^2$ and $\sigma_{4}^2$.

 Similarly, based on the derivations of Eq.~\ref{eq:multitask_detail} and \ref{eq:approx}, we can also derive the loss $L^{arm}$ for anchor refinement module (ARM). Therefore, the final overall loss for balanced multi-task training is as Eq.~\ref{eq:multitask} in original paper.
 In practice, for the stable when optimizing the overall loss, the trainable parameters consist of network parameters $\phi$ and logarithmic form of the four uncertainty parameters, namely \{$\log\sigma_1^2, \log\sigma_2^2,\log\sigma_3^2,\log\sigma_4^2$\}.
\section{More visualizations}
\begin{figure*}[t]
\begin{minipage}[t]{0.195\linewidth}
  \centering
  \centerline{\includegraphics[width=3.5cm, height=4cm]{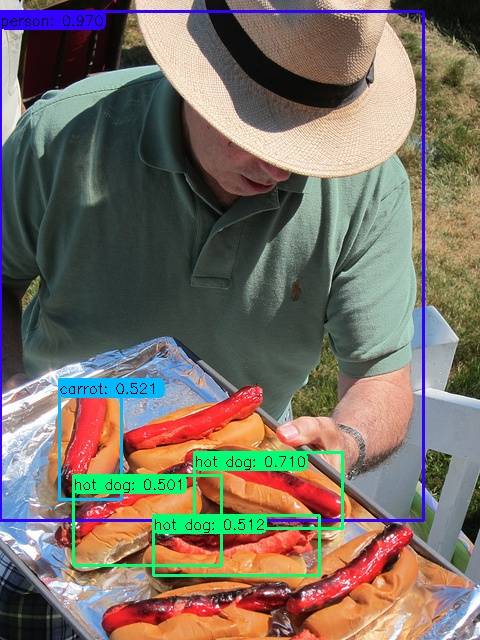}}
\end{minipage}
\hfill
\begin{minipage}[t]{0.195\linewidth}
  \centering
  \centerline{\includegraphics[width=3.5cm, height=4cm]{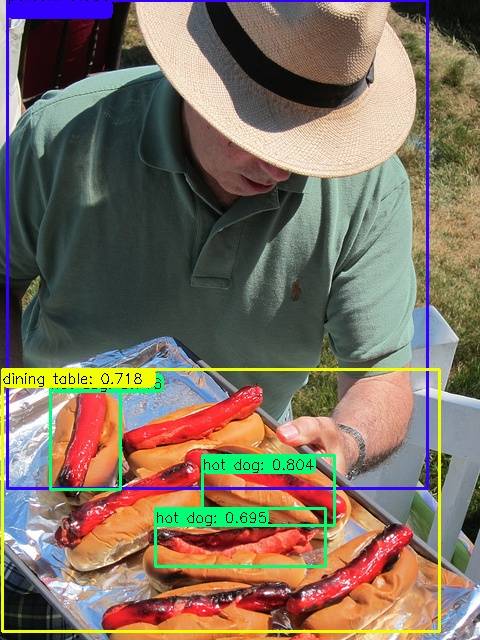}}
\end{minipage}
\hfill
\begin{minipage}[t]{0.195\linewidth}
  \centering
  \centerline{\includegraphics[width=3.5cm, height=4cm]{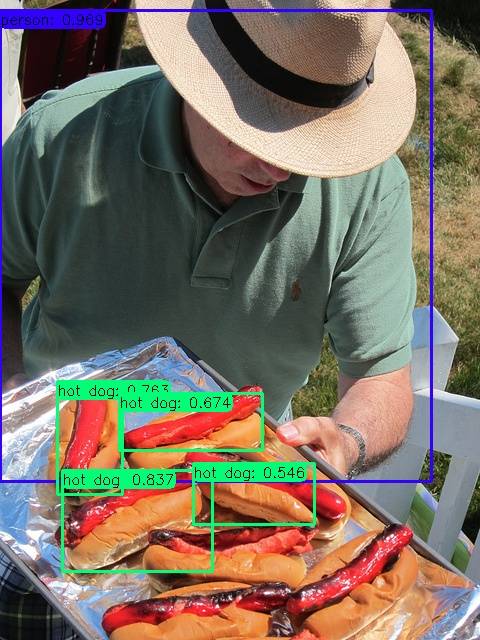}}
\end{minipage}
\hfill
\begin{minipage}[t]{0.195\linewidth}
  \centering
  \centerline{\includegraphics[width=3.5cm, height=4cm]{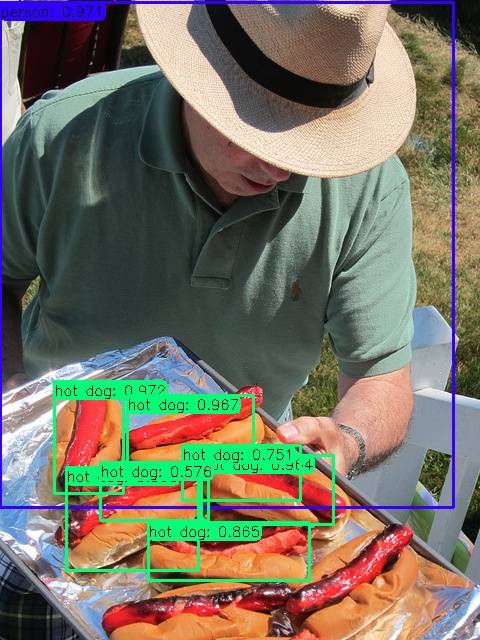}}
\end{minipage}
\hfill
\begin{minipage}[t]{0.195\linewidth}
  \centering
  \centerline{\includegraphics[width=3.5cm, height=4cm]{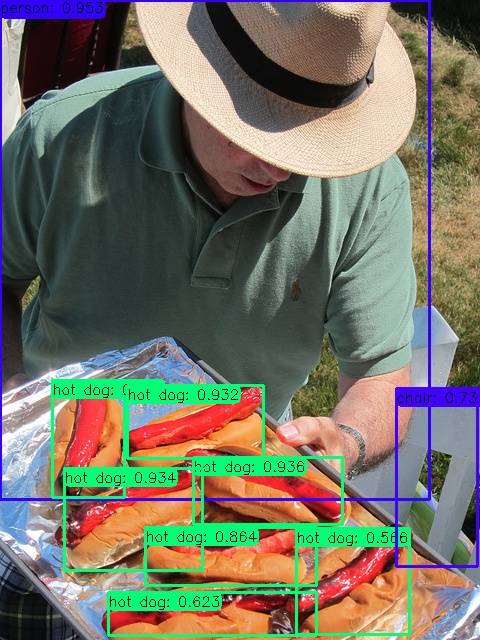}}
\end{minipage}\\
\begin{minipage}[t]{0.195\linewidth}
  \centering
  \centerline{\includegraphics[width=3.5cm]{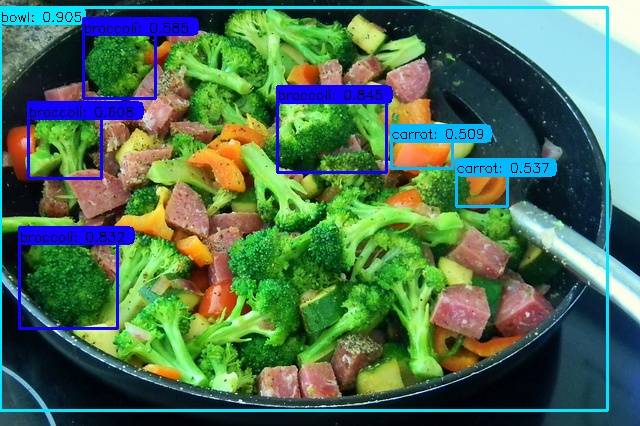}}
\end{minipage}
\hfill
\begin{minipage}[t]{0.195\linewidth}
  \centering
  \centerline{\includegraphics[width=3.5cm]{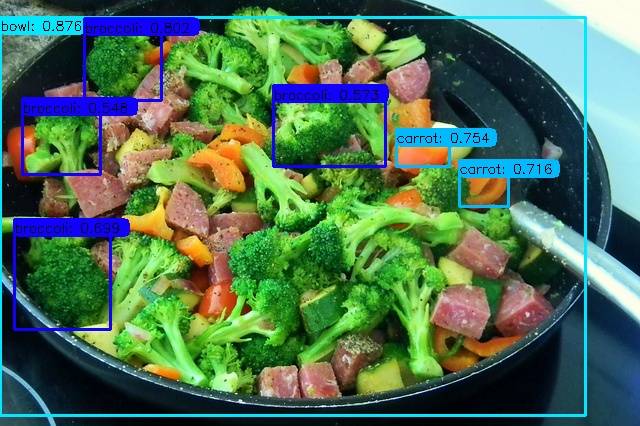}}
\end{minipage}
\hfill
\begin{minipage}[t]{0.195\linewidth}
  \centering
  \centerline{\includegraphics[width=3.5cm]{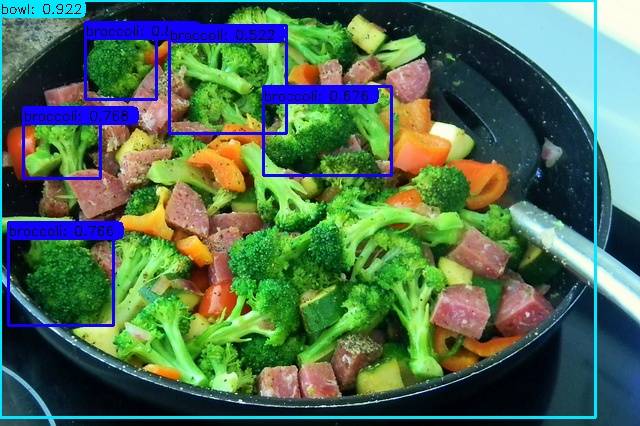}}
\end{minipage}
\hfill
\begin{minipage}[t]{0.195\linewidth}
  \centering
  \centerline{\includegraphics[width=3.5cm]{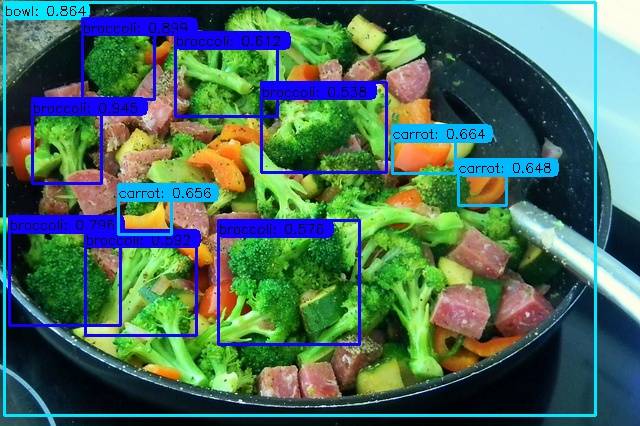}}
\end{minipage}
\hfill
\begin{minipage}[t]{0.195\linewidth}
  \centering
  \centerline{\includegraphics[width=3.5cm]{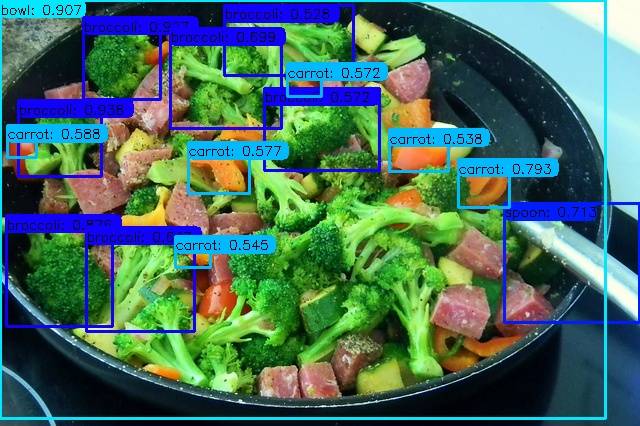}}
\end{minipage}\\
\begin{minipage}[t]{0.195\linewidth}
  \centering
  \centerline{\includegraphics[width=3.5cm, height=4.5cm]{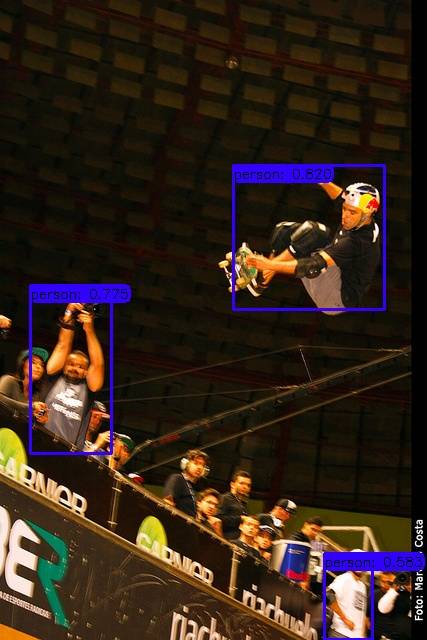}}
\end{minipage}
\hfill
\begin{minipage}[t]{0.195\linewidth}
  \centering
  \centerline{\includegraphics[width=3.5cm, height=4.5cm]{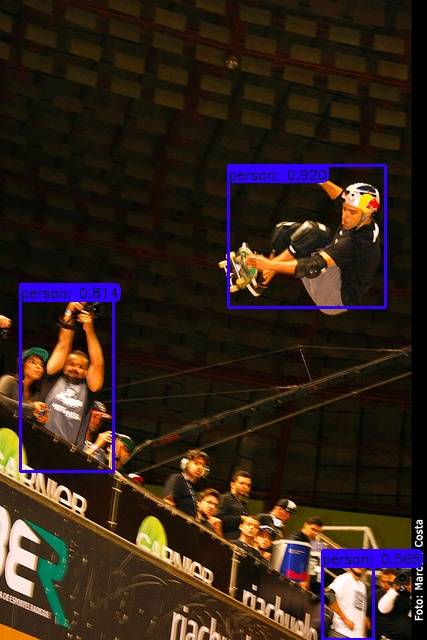}}
\end{minipage}
\hfill
\begin{minipage}[t]{0.195\linewidth}
  \centering
  \centerline{\includegraphics[width=3.5cm, height=4.5cm]{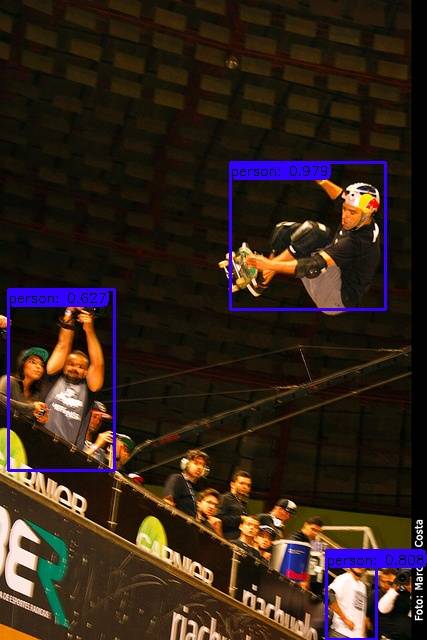}}
\end{minipage}
\hfill
\begin{minipage}[t]{0.195\linewidth}
  \centering
  \centerline{\includegraphics[width=3.5cm, height=4.5cm]{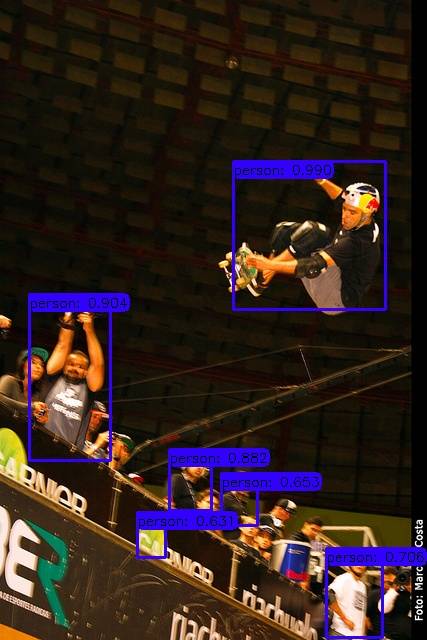}}
\end{minipage}
\hfill
\begin{minipage}[t]{0.195\linewidth}
  \centering
  \centerline{\includegraphics[width=3.5cm, height=4.5cm]{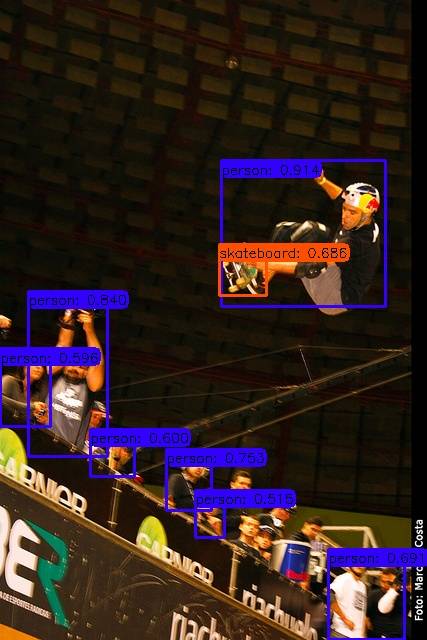}}
\end{minipage}\\
\begin{minipage}[t]{0.195\linewidth}
  \centering
  \centerline{\includegraphics[width=3.5cm]{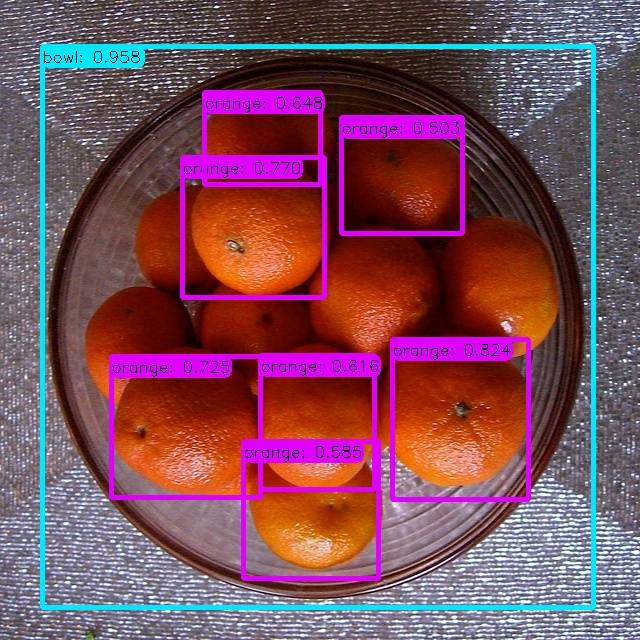}}
\end{minipage}
\hfill
\begin{minipage}[t]{0.195\linewidth}
  \centering
  \centerline{\includegraphics[width=3.5cm]{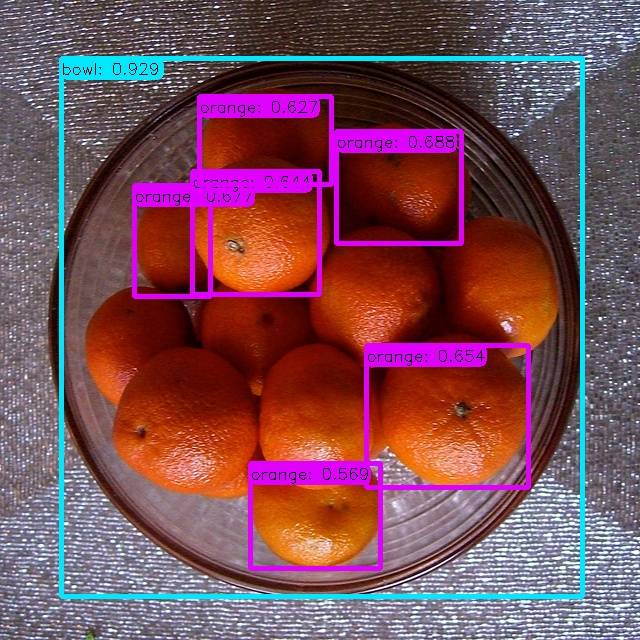}}
\end{minipage}
\hfill
\begin{minipage}[t]{0.195\linewidth}
  \centering
  \centerline{\includegraphics[width=3.5cm]{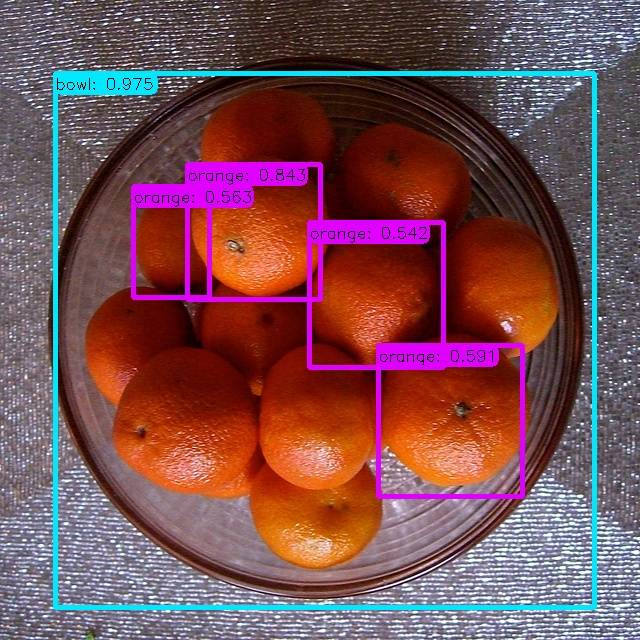}}
\end{minipage}
\hfill
\begin{minipage}[t]{0.195\linewidth}
  \centering
  \centerline{\includegraphics[width=3.5cm]{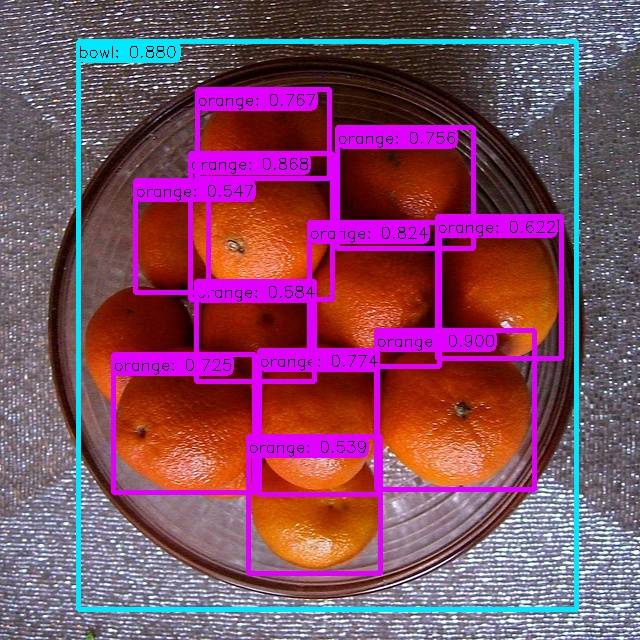}}
\end{minipage}
\hfill
\begin{minipage}[t]{0.195\linewidth}
  \centering
  \centerline{\includegraphics[width=3.5cm]{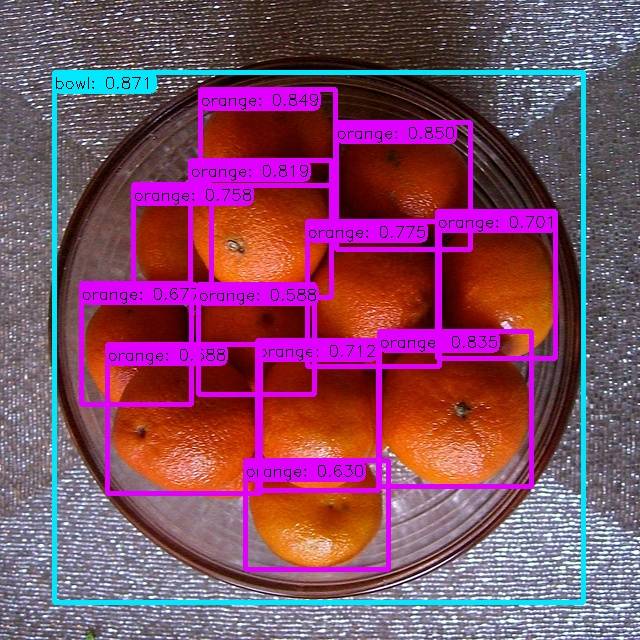}}
\end{minipage}\\
\begin{minipage}[t]{0.195\linewidth}
  \centering
  \centerline{\includegraphics[width=3.5cm]{000000013923_detect_SSDLite}}
\end{minipage}
\hfill
\begin{minipage}[t]{0.195\linewidth}
  \centering
  \centerline{\includegraphics[width=3.5cm]{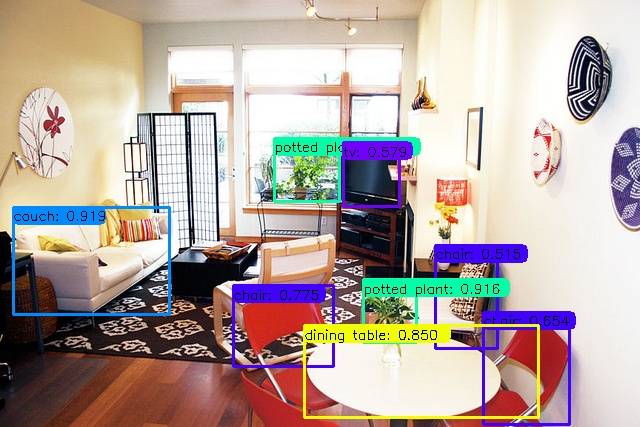}}
\end{minipage}
\hfill
\begin{minipage}[t]{0.195\linewidth}
  \centering
  \centerline{\includegraphics[width=3.5cm]{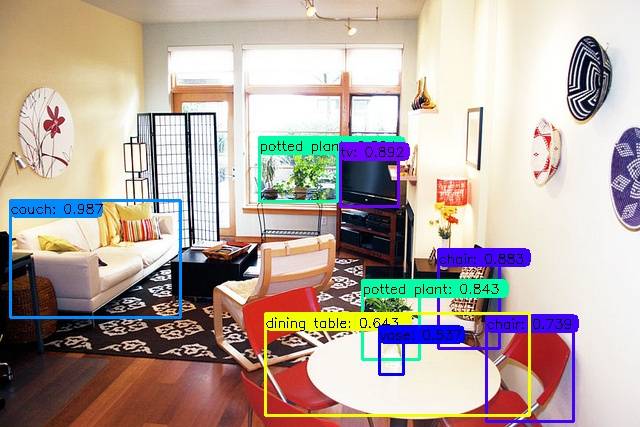}}
\end{minipage}
\hfill
\begin{minipage}[t]{0.195\linewidth}
  \centering
  \centerline{\includegraphics[width=3.5cm]{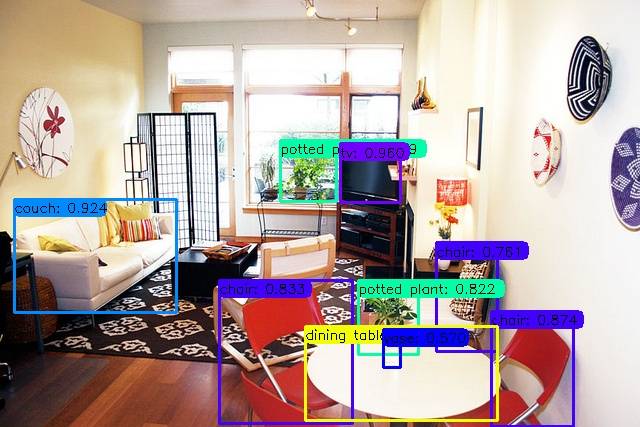}}
\end{minipage}
\hfill
\begin{minipage}[t]{0.195\linewidth}
  \centering
  \centerline{\includegraphics[width=3.5cm]{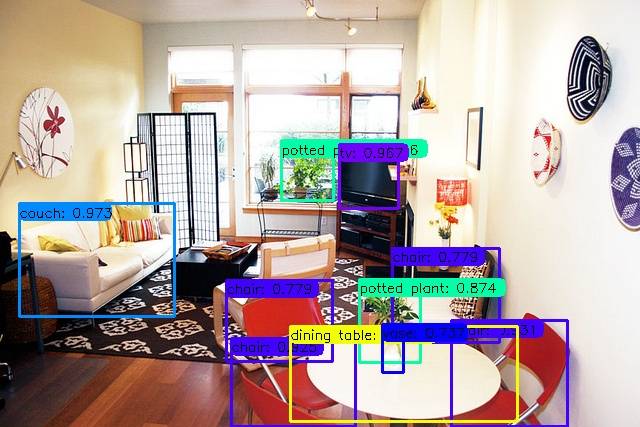}}
\end{minipage}\\
\begin{minipage}[t]{0.195\linewidth}
  \centering
  \centerline{\includegraphics[width=3.5cm, height=4.5cm]{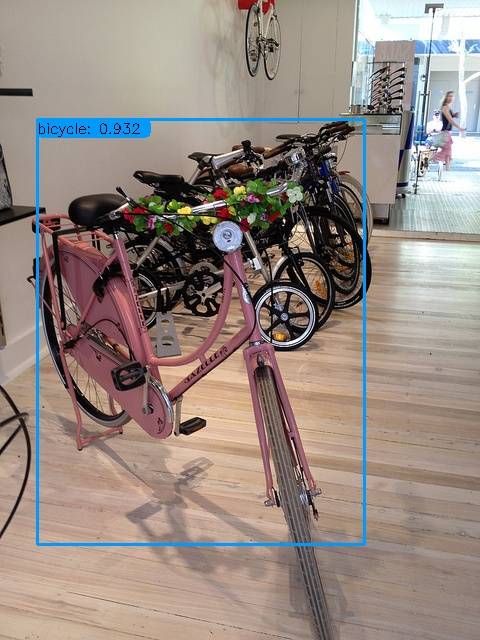}}
  \centerline{\small{(a) SSDLite+MobileNet}}
\end{minipage}
\hfill
\begin{minipage}[t]{0.195\linewidth}
  \centering
  \centerline{\includegraphics[width=3.5cm, height=4.5cm]{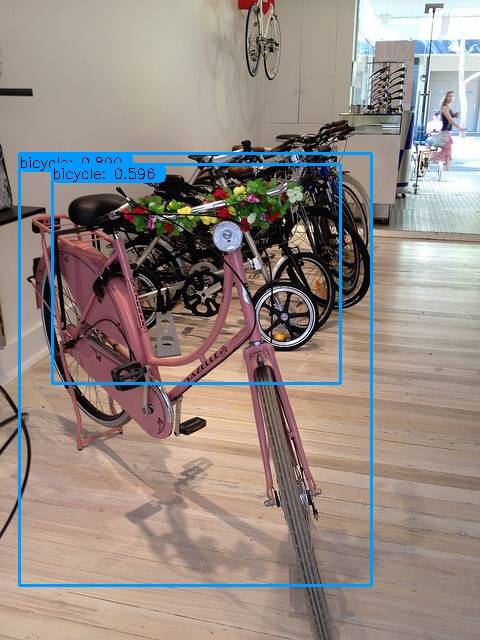}}
  \centerline{\small{(b) RefineDet+ShuffleNetV2}}
\end{minipage}
\hfill
\begin{minipage}[t]{0.195\linewidth}
  \centering
  \centerline{\includegraphics[width=3.5cm, height=4.5cm]{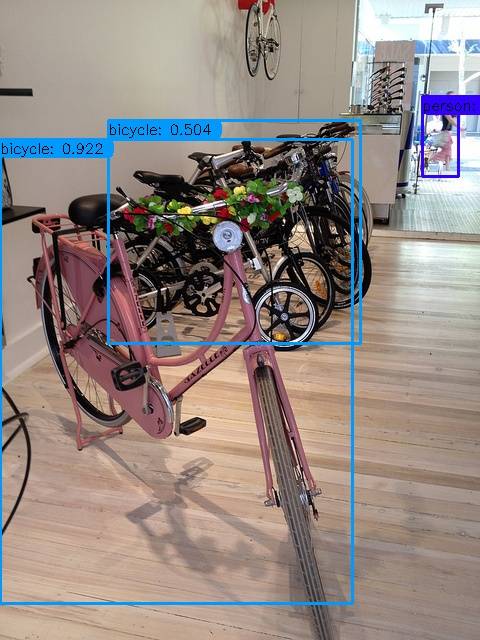}}
  \centerline{\small{(c) SSD+VGG}}
\end{minipage}
\hfill
\begin{minipage}[t]{0.195\linewidth}
  \centering
  \centerline{\includegraphics[width=3.5cm, height=4.5cm]{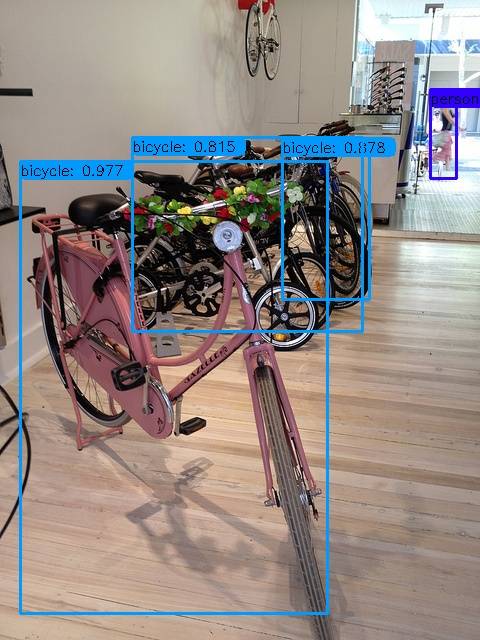}}
  \centerline{\small{(d) RefineDetLite}}
\end{minipage}
\hfill
\begin{minipage}[t]{0.195\linewidth}
  \centering
  \centerline{\includegraphics[width=3.5cm, height=4.5cm]{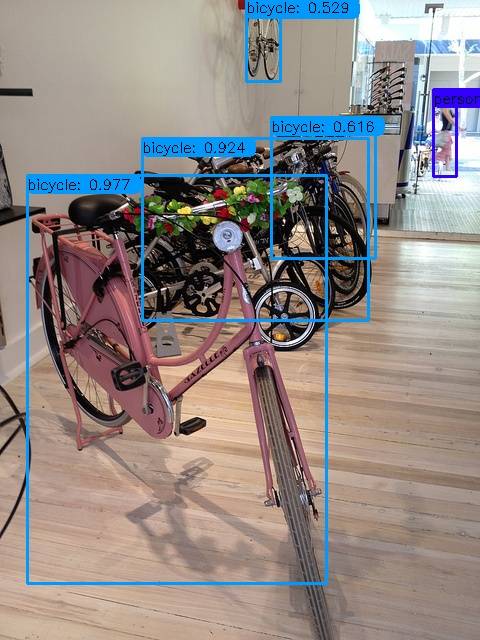}}
  \centerline{(e) RefineDetLite++}
\end{minipage}
\caption{Visualization comparisons among different models. This groups of images show that RefinDetLite can detect more small objects. Furthermore, RefineDetLite++ can refine the results more accurately. Best viewed by zooming in.}
\label{fig:visualization_smallobjects}
\end{figure*}

\begin{figure*}[t]
\begin{minipage}[t]{0.195\linewidth}
  \centering
  \centerline{\includegraphics[width=3.5cm, height=3.7cm]{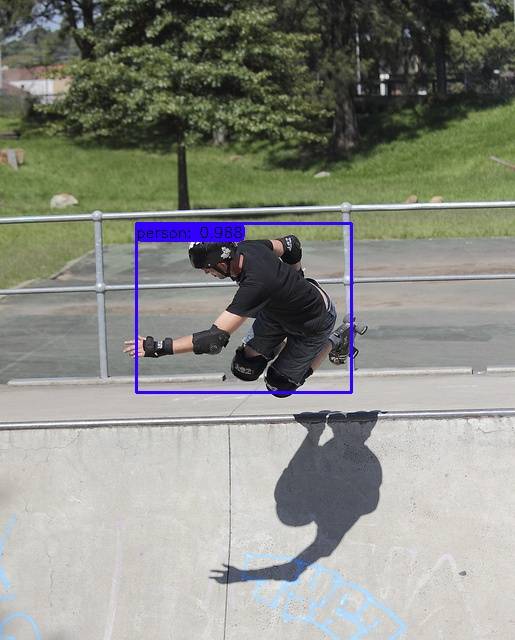}}
\end{minipage}
\hfill
\begin{minipage}[t]{0.195\linewidth}
  \centering
  \centerline{\includegraphics[width=3.5cm, height=3.7cm]{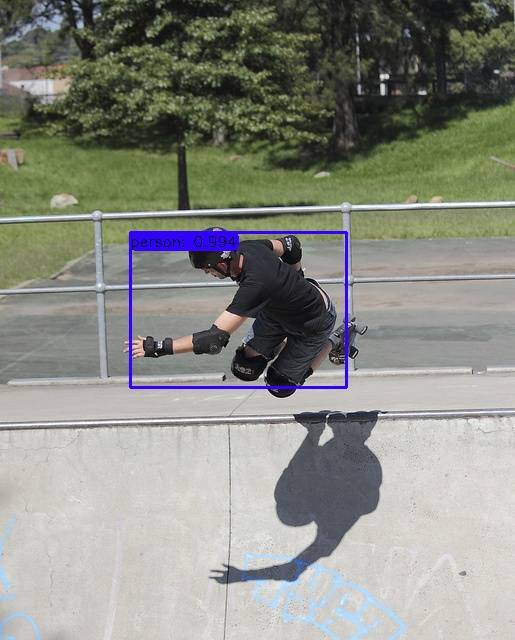}}
\end{minipage}
\hfill
\begin{minipage}[t]{0.195\linewidth}
  \centering
  \centerline{\includegraphics[width=3.5cm, height=3.7cm]{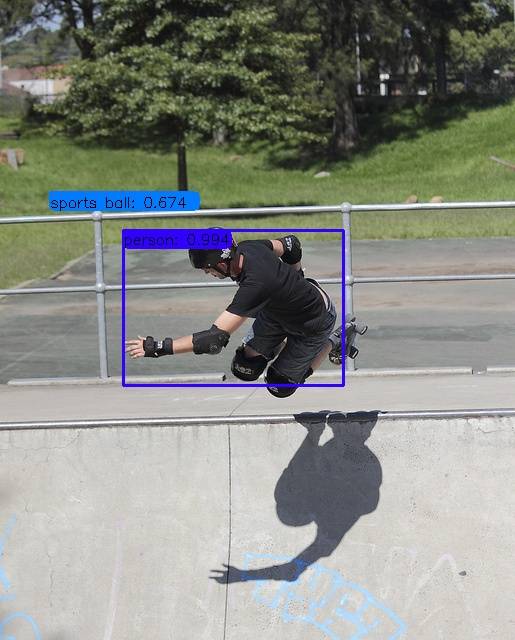}}
\end{minipage}
\hfill
\begin{minipage}[t]{0.195\linewidth}
  \centering
  \centerline{\includegraphics[width=3.5cm, height=3.7cm]{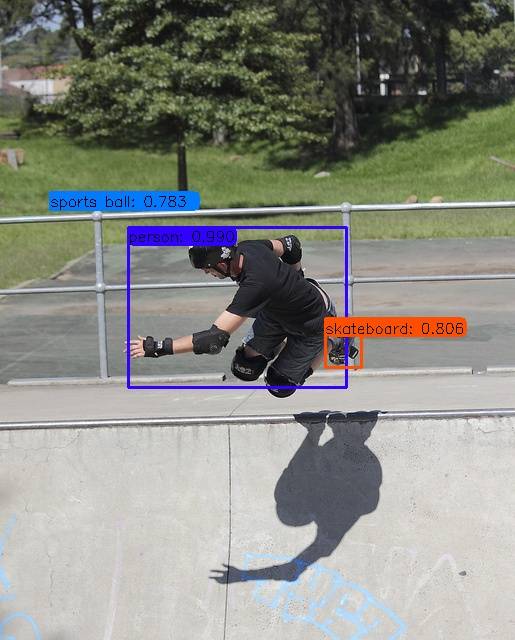}}
\end{minipage}
\hfill
\begin{minipage}[t]{0.195\linewidth}
  \centering
  \centerline{\includegraphics[width=3.5cm, height=3.7cm]{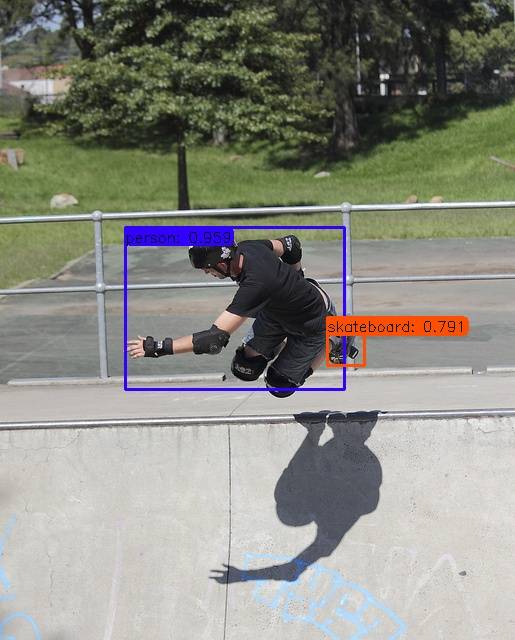}}
\end{minipage}\\
\begin{minipage}[t]{0.195\linewidth}
  \centering
  \centerline{\includegraphics[width=3.5cm]{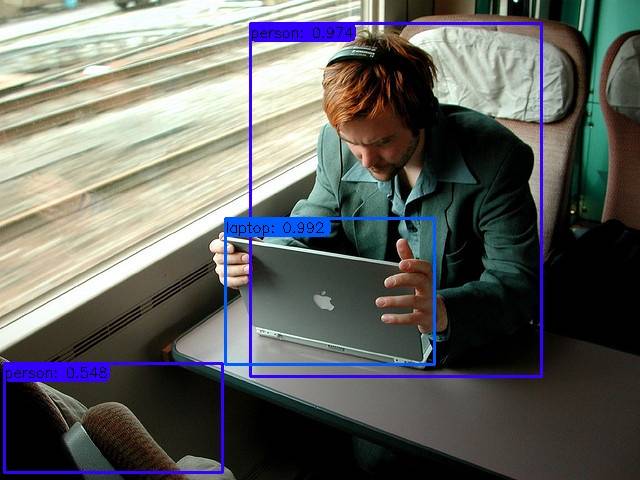}}
\end{minipage}
\hfill
\begin{minipage}[t]{0.195\linewidth}
  \centering
  \centerline{\includegraphics[width=3.5cm]{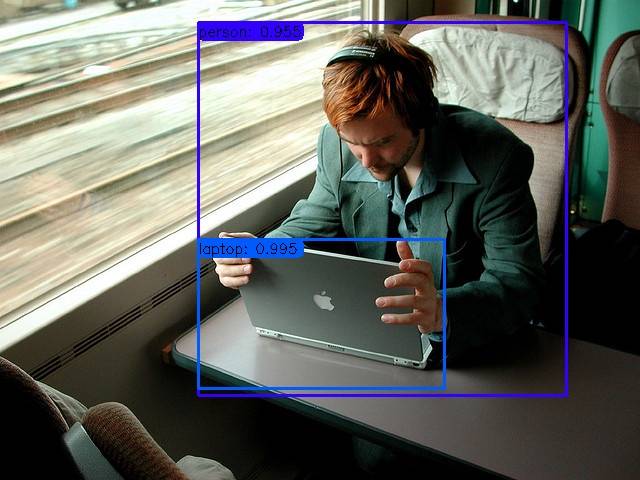}}
\end{minipage}
\hfill
\begin{minipage}[t]{0.195\linewidth}
  \centering
  \centerline{\includegraphics[width=3.5cm]{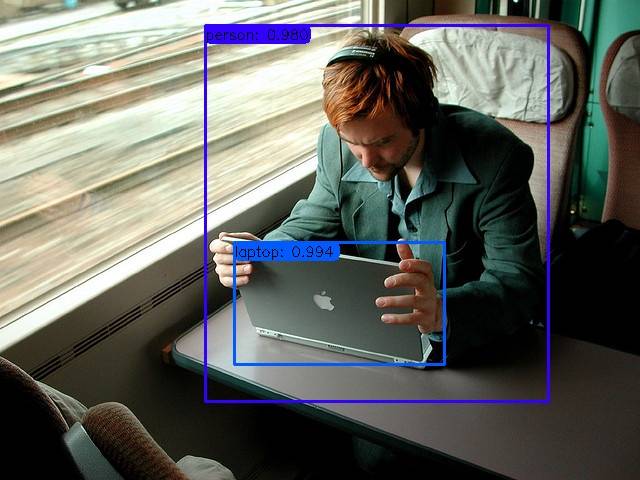}}
\end{minipage}
\hfill
\begin{minipage}[t]{0.195\linewidth}
  \centering
  \centerline{\includegraphics[width=3.5cm]{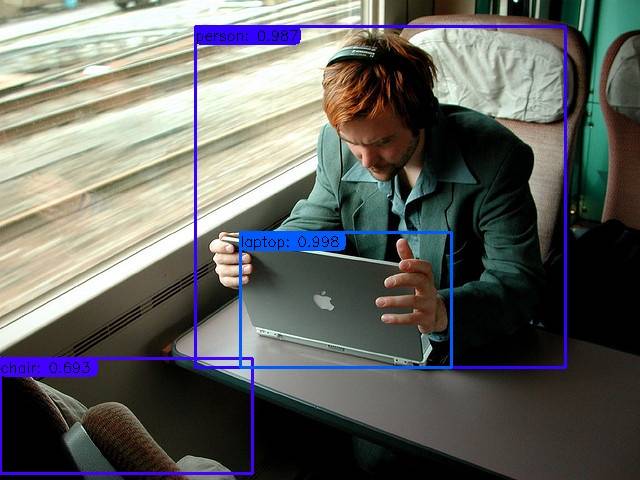}}
\end{minipage}
\hfill
\begin{minipage}[t]{0.195\linewidth}
  \centering
  \centerline{\includegraphics[width=3.5cm]{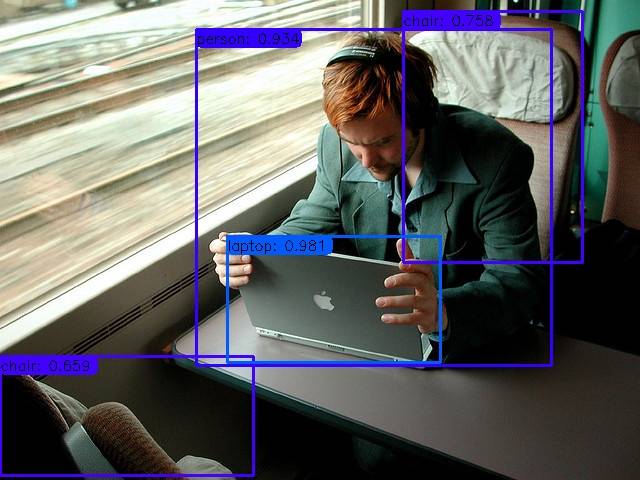}}
\end{minipage}\\
\begin{minipage}[t]{0.195\linewidth}
  \centering
  \centerline{\includegraphics[width=3.5cm, height=3.8cm]{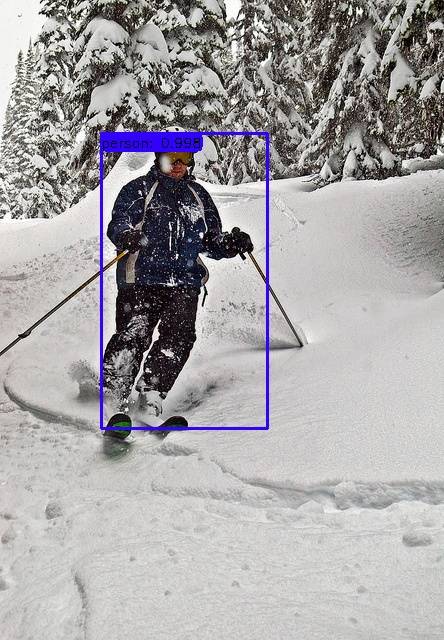}}
\end{minipage}
\hfill
\begin{minipage}[t]{0.195\linewidth}
  \centering
  \centerline{\includegraphics[width=3.5cm, height=3.8cm]{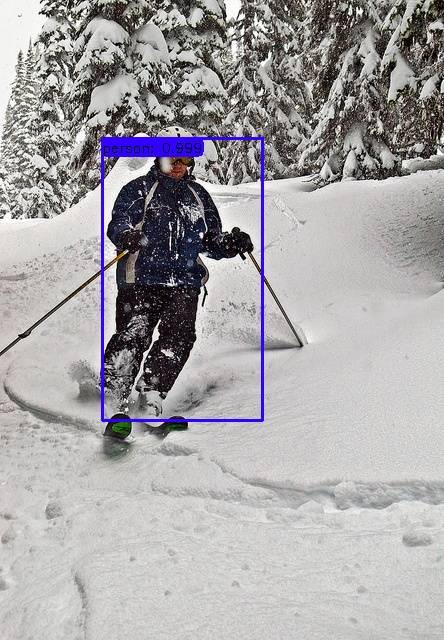}}
\end{minipage}
\hfill
\begin{minipage}[t]{0.195\linewidth}
  \centering
  \centerline{\includegraphics[width=3.5cm, height=3.8cm]{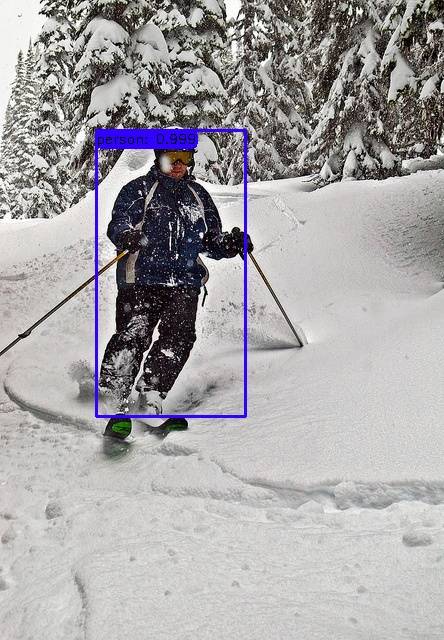}}
\end{minipage}
\hfill
\begin{minipage}[t]{0.195\linewidth}
  \centering
  \centerline{\includegraphics[width=3.5cm, height=3.8cm]{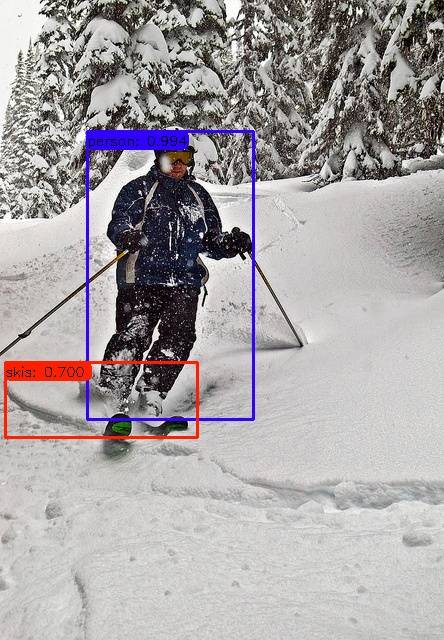}}
\end{minipage}
\hfill
\begin{minipage}[t]{0.195\linewidth}
  \centering
  \centerline{\includegraphics[width=3.5cm, height=3.8cm]{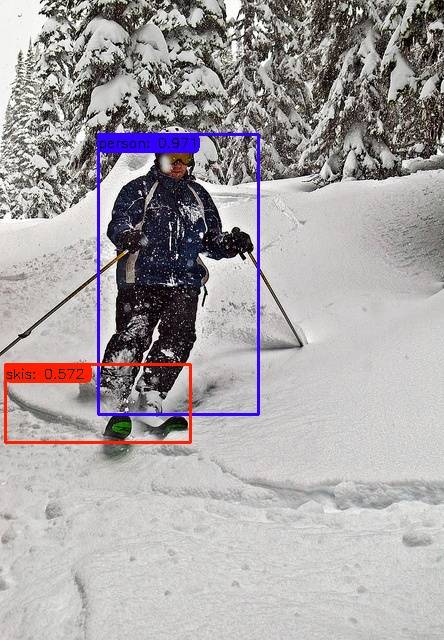}}
\end{minipage}\\
\begin{minipage}[t]{0.195\linewidth}
  \centering
  \centerline{\includegraphics[width=3.5cm]{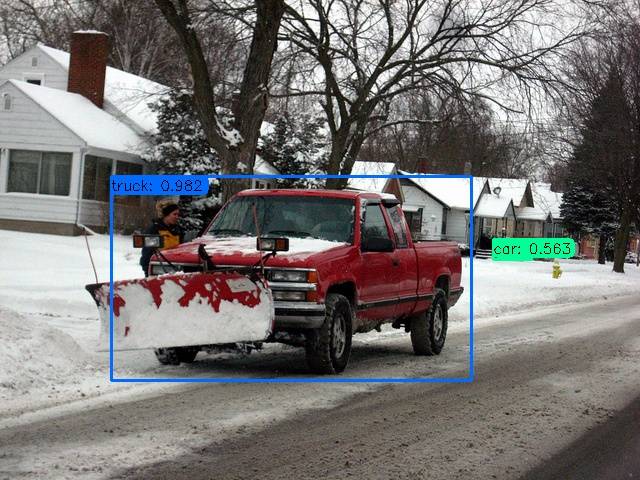}}
\end{minipage}
\hfill
\begin{minipage}[t]{0.195\linewidth}
  \centering
  \centerline{\includegraphics[width=3.5cm]{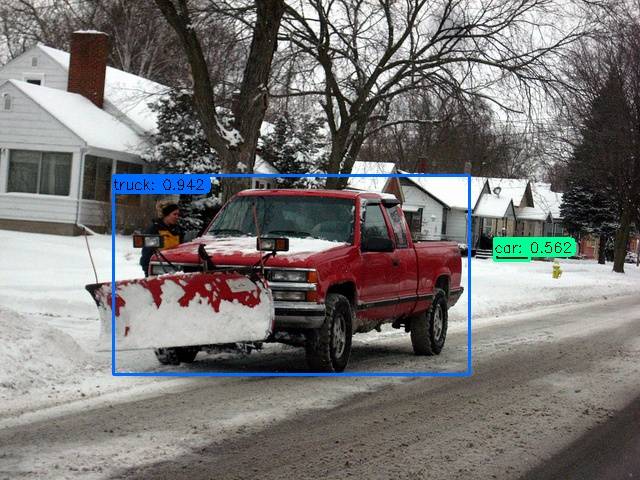}}
\end{minipage}
\hfill
\begin{minipage}[t]{0.195\linewidth}
  \centering
  \centerline{\includegraphics[width=3.5cm]{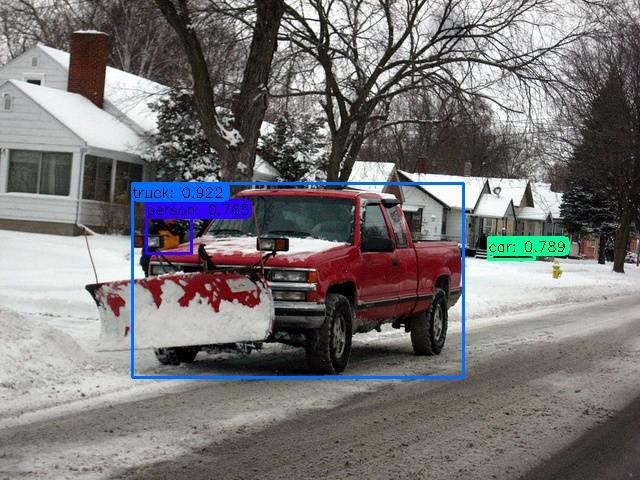}}
\end{minipage}
\hfill
\begin{minipage}[t]{0.195\linewidth}
  \centering
  \centerline{\includegraphics[width=3.5cm]{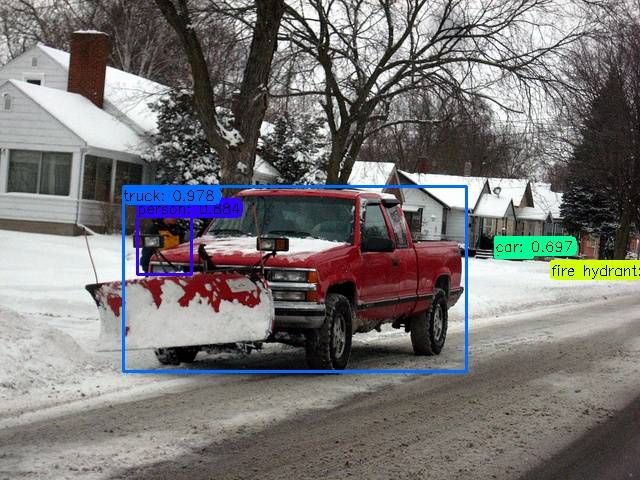}}
\end{minipage}
\hfill
\begin{minipage}[t]{0.195\linewidth}
  \centering
  \centerline{\includegraphics[width=3.5cm]{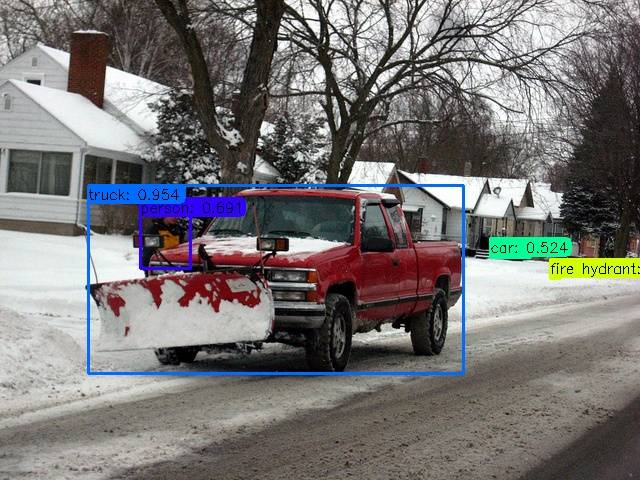}}
\end{minipage}\\
\begin{minipage}[t]{0.195\linewidth}
  \centering
  \centerline{\includegraphics[width=3.5cm]{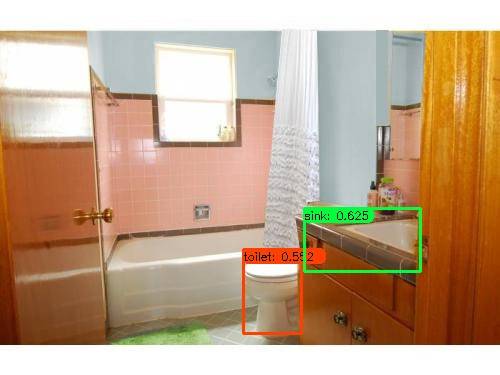}}
\end{minipage}
\hfill
\begin{minipage}[t]{0.195\linewidth}
  \centering
  \centerline{\includegraphics[width=3.5cm]{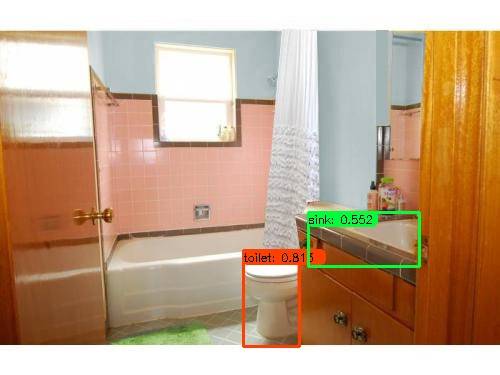}}
\end{minipage}
\hfill
\begin{minipage}[t]{0.195\linewidth}
  \centering
  \centerline{\includegraphics[width=3.5cm]{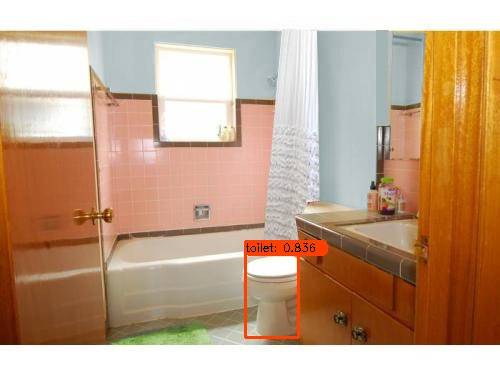}}
\end{minipage}
\hfill
\begin{minipage}[t]{0.195\linewidth}
  \centering
  \centerline{\includegraphics[width=3.5cm]{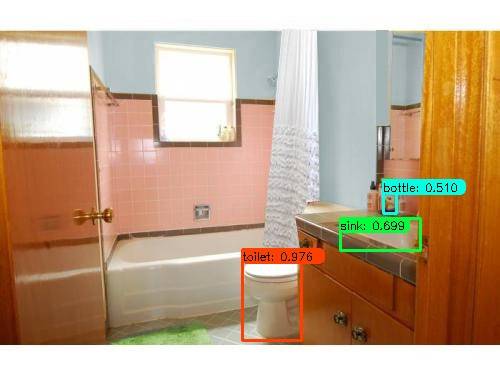}}
\end{minipage}
\hfill
\begin{minipage}[t]{0.195\linewidth}
  \centering
  \centerline{\includegraphics[width=3.5cm]{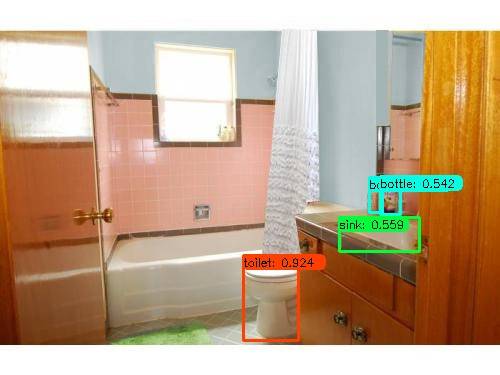}}
\end{minipage}\\
\begin{minipage}[t]{0.195\linewidth}
  \centering
  \centerline{\includegraphics[width=3.5cm, height=3.8cm]{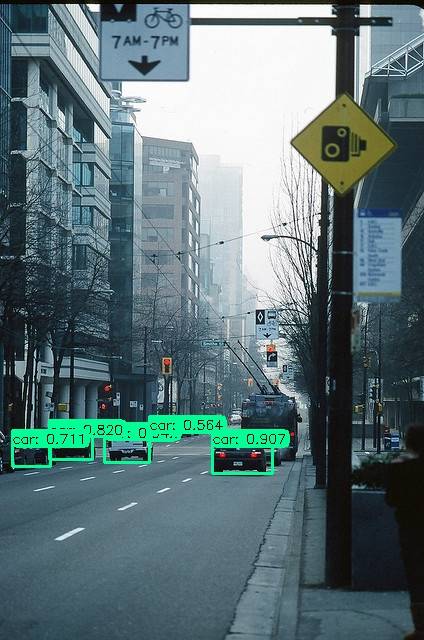}}
\end{minipage}
\hfill
\begin{minipage}[t]{0.195\linewidth}
  \centering
  \centerline{\includegraphics[width=3.5cm, height=3.8cm]{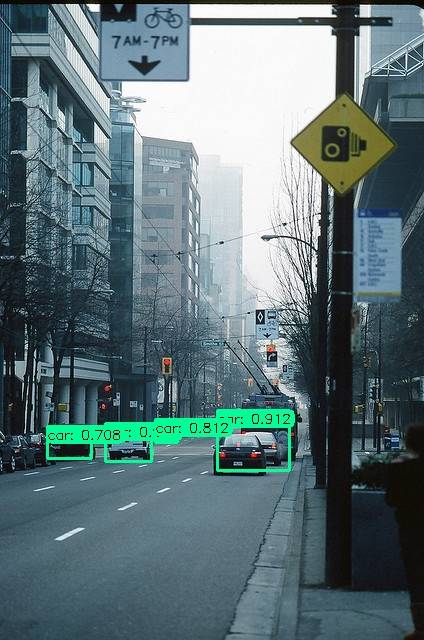}}
\end{minipage}
\hfill
\begin{minipage}[t]{0.195\linewidth}
  \centering
  \centerline{\includegraphics[width=3.5cm, height=3.8cm]{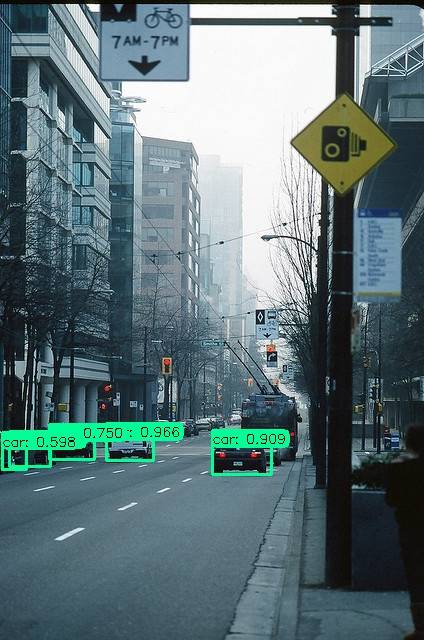}}
\end{minipage}
\hfill
\begin{minipage}[t]{0.195\linewidth}
  \centering
  \centerline{\includegraphics[width=3.5cm, height=3.8cm]{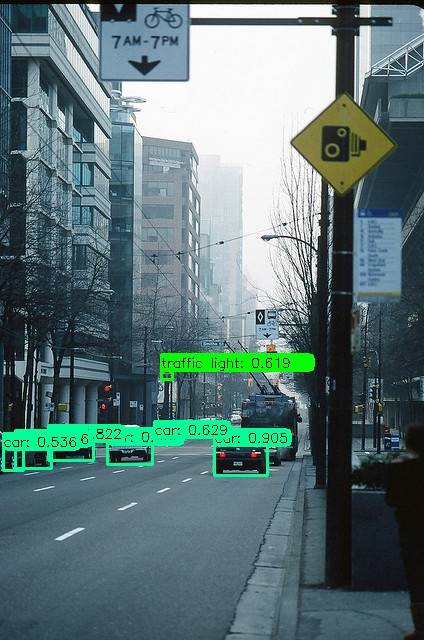}}
\end{minipage}
\hfill
\begin{minipage}[t]{0.195\linewidth}
  \centering
  \centerline{\includegraphics[width=3.5cm, height=3.8cm]{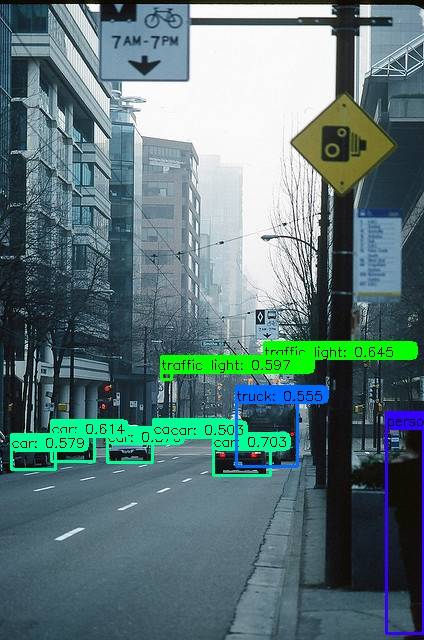}}
\end{minipage}\\
\begin{minipage}[t]{0.195\linewidth}
  \centering
  \centerline{\includegraphics[width=3.5cm]{000000044260_detect_SSDLite}}
  \centerline{\small{(a) SSDLite+MobileNet}}
\end{minipage}
\hfill
\begin{minipage}[t]{0.195\linewidth}
  \centering
  \centerline{\includegraphics[width=3.5cm]{000000044260_detect_shufflenetv2}}
  \centerline{\small{(b) RefineDet+ShuffleNetV2}}
\end{minipage}
\hfill
\begin{minipage}[t]{0.195\linewidth}
  \centering
  \centerline{\includegraphics[width=3.5cm]{000000044260_detect_SSD}}
  \centerline{\small{(c) SSD+VGG}}
\end{minipage}
\hfill
\begin{minipage}[t]{0.195\linewidth}
  \centering
  \centerline{\includegraphics[width=3.5cm]{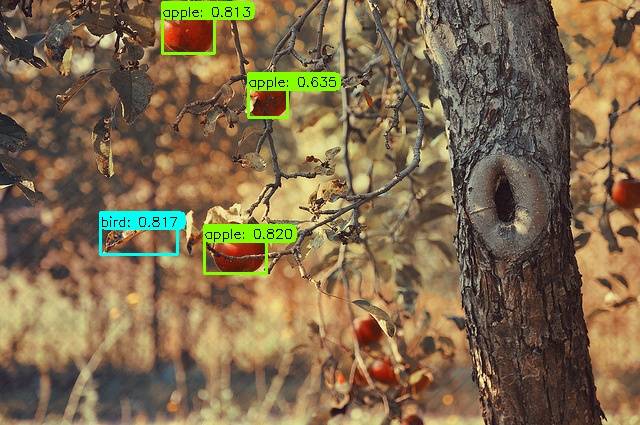}}
  \centerline{\small{(d) RefineDetLite}}
\end{minipage}
\hfill
\begin{minipage}[t]{0.195\linewidth}
  \centering
  \centerline{\includegraphics[width=3.5cm]{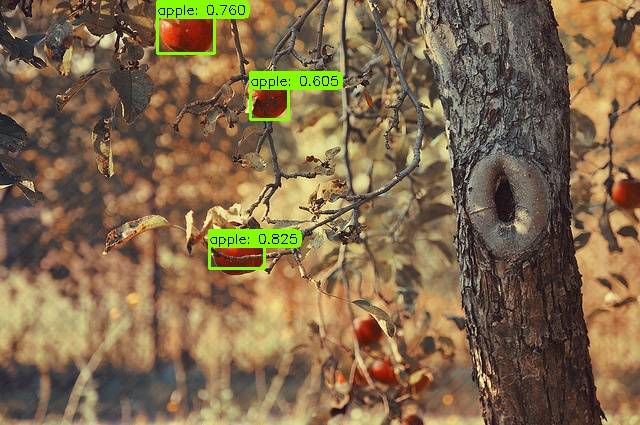}}
  \centerline{(e) RefineDetLite++}
\end{minipage}
\caption{Visualization comparisons among different models. This groups of images show that RefinDetLite can detect more kinds of objects, especially small objects. Furthermore, RefineDetLite++ can refine the results more accurately. Best viewed by zooming in.}
\label{fig:visualization_moreobjectsclasses}
\end{figure*}

\begin{figure*}[t]
\begin{minipage}[t]{0.195\linewidth}
  \centering
  \centerline{\includegraphics[width=3.5cm]{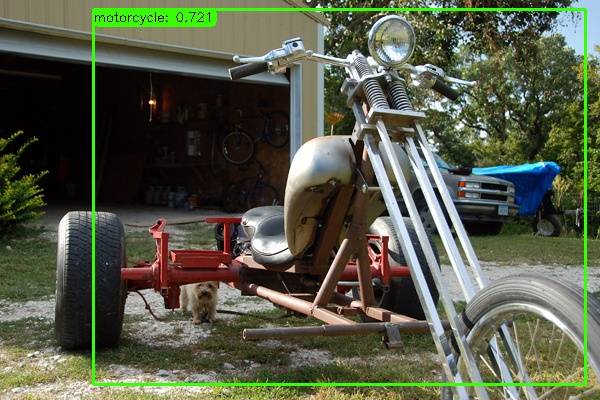}}
\end{minipage}
\hfill
\begin{minipage}[t]{0.195\linewidth}
  \centering
  \centerline{\includegraphics[width=3.5cm]{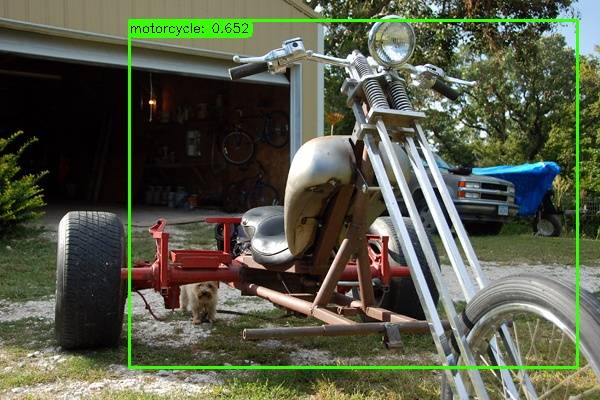}}
\end{minipage}
\hfill
\begin{minipage}[t]{0.195\linewidth}
  \centering
  \centerline{\includegraphics[width=3.5cm]{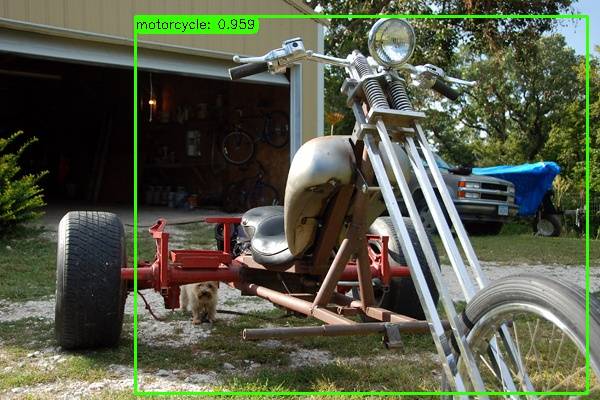}}
\end{minipage}
\hfill
\begin{minipage}[t]{0.195\linewidth}
  \centering
  \centerline{\includegraphics[width=3.5cm]{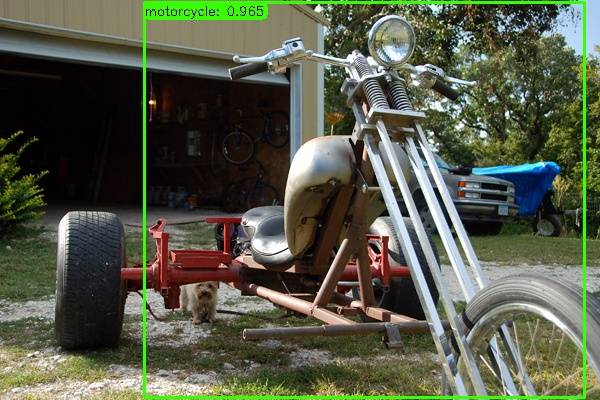}}
\end{minipage}
\hfill
\begin{minipage}[t]{0.195\linewidth}
  \centering
  \centerline{\includegraphics[width=3.5cm]{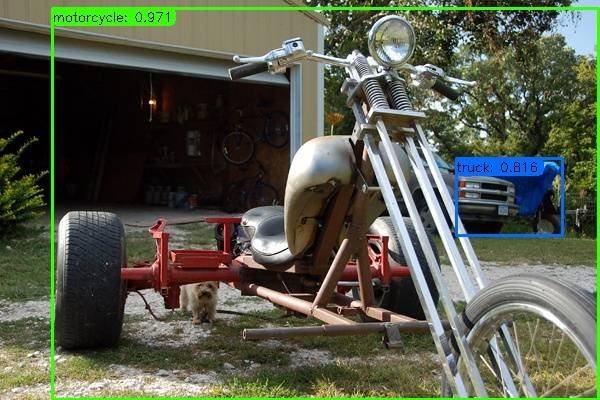}}
\end{minipage}\\
\begin{minipage}[t]{0.195\linewidth}
  \centering
  \centerline{\includegraphics[width=3.5cm]{000000009448_detect_SSDLite}}
\end{minipage}
\hfill
\begin{minipage}[t]{0.195\linewidth}
  \centering
  \centerline{\includegraphics[width=3.5cm]{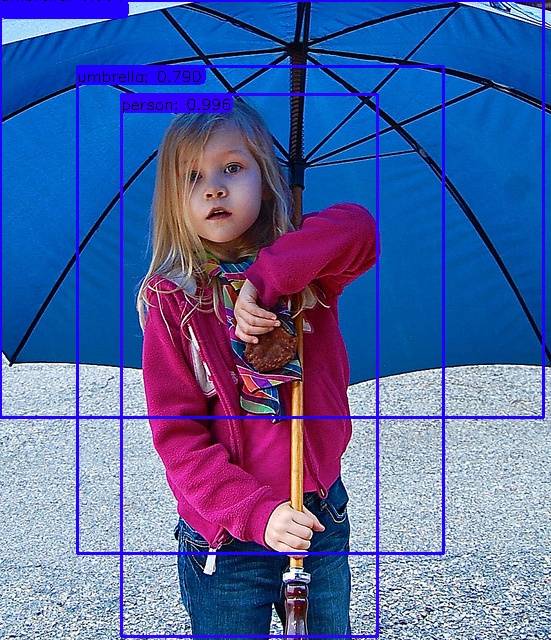}}
\end{minipage}
\hfill
\begin{minipage}[t]{0.195\linewidth}
  \centering
  \centerline{\includegraphics[width=3.5cm]{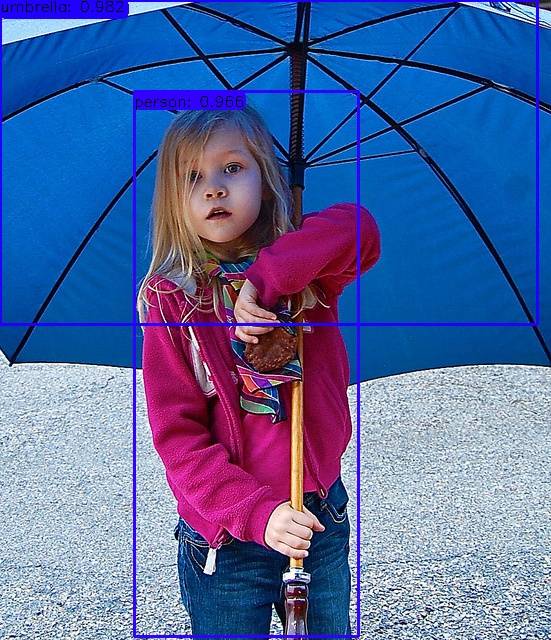}}
\end{minipage}
\hfill
\begin{minipage}[t]{0.195\linewidth}
  \centering
  \centerline{\includegraphics[width=3.5cm]{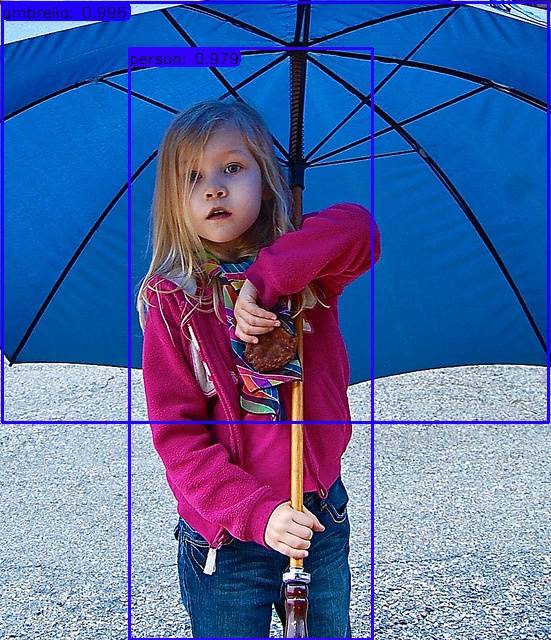}}
\end{minipage}
\hfill
\begin{minipage}[t]{0.195\linewidth}
  \centering
  \centerline{\includegraphics[width=3.5cm]{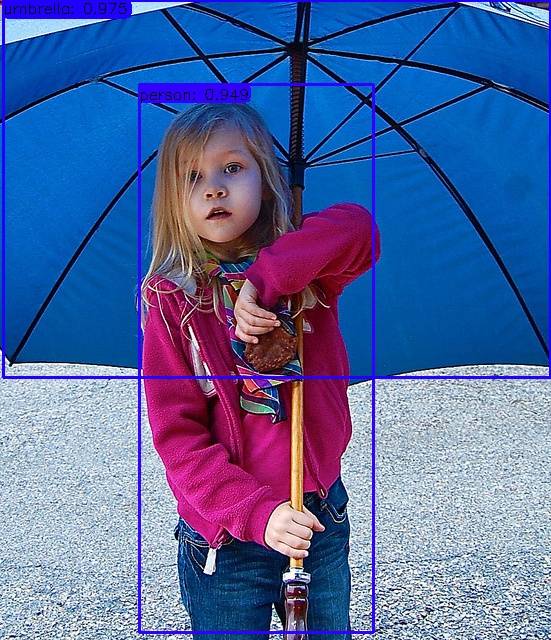}}
\end{minipage}\\
\begin{minipage}[t]{0.195\linewidth}
  \centering
  \centerline{\includegraphics[width=3.5cm]{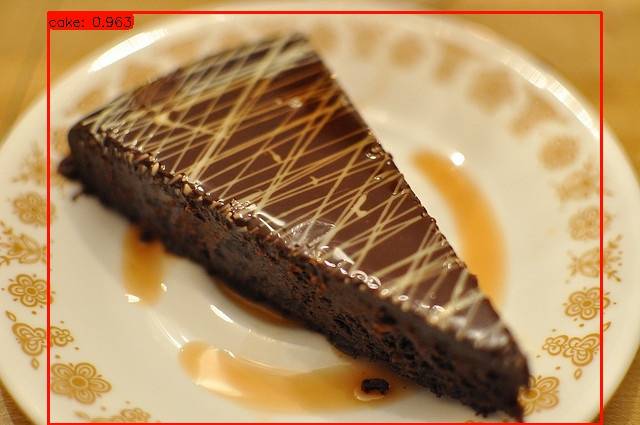}}
\end{minipage}
\hfill
\begin{minipage}[t]{0.195\linewidth}
  \centering
  \centerline{\includegraphics[width=3.5cm]{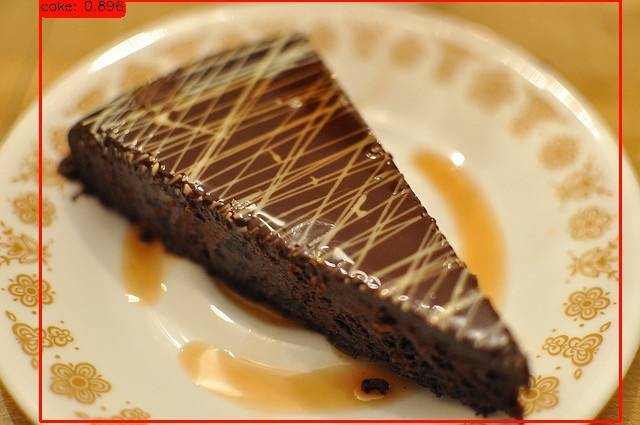}}
\end{minipage}
\hfill
\begin{minipage}[t]{0.195\linewidth}
  \centering
  \centerline{\includegraphics[width=3.5cm]{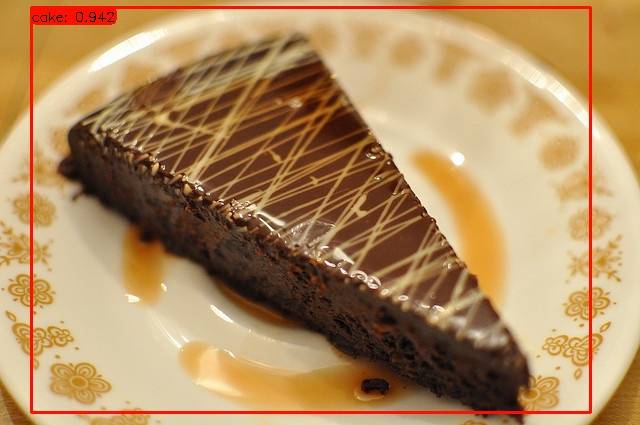}}
\end{minipage}
\hfill
\begin{minipage}[t]{0.195\linewidth}
  \centering
  \centerline{\includegraphics[width=3.5cm]{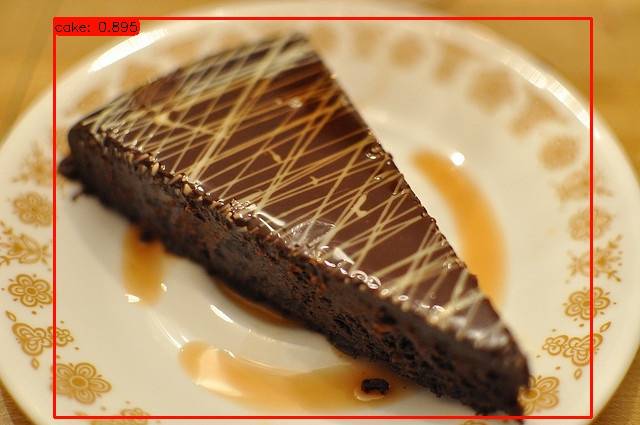}}
\end{minipage}
\hfill
\begin{minipage}[t]{0.195\linewidth}
  \centering
  \centerline{\includegraphics[width=3.5cm]{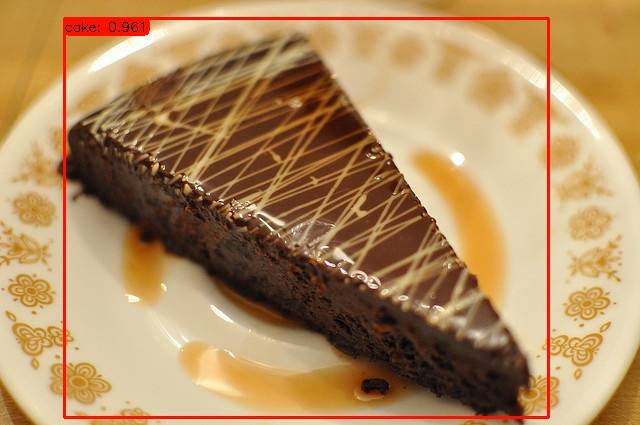}}
\end{minipage}\\
\begin{minipage}[t]{0.195\linewidth}
  \centering
  \centerline{\includegraphics[width=3.5cm]{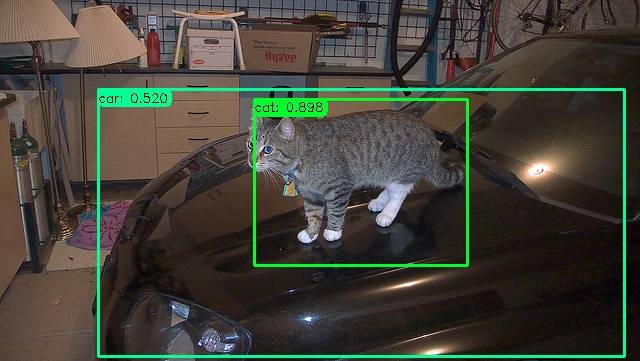}}
\end{minipage}
\hfill
\begin{minipage}[t]{0.195\linewidth}
  \centering
  \centerline{\includegraphics[width=3.5cm]{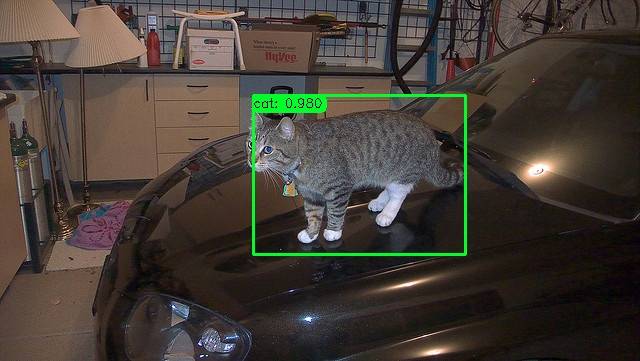}}
\end{minipage}
\hfill
\begin{minipage}[t]{0.195\linewidth}
  \centering
  \centerline{\includegraphics[width=3.5cm]{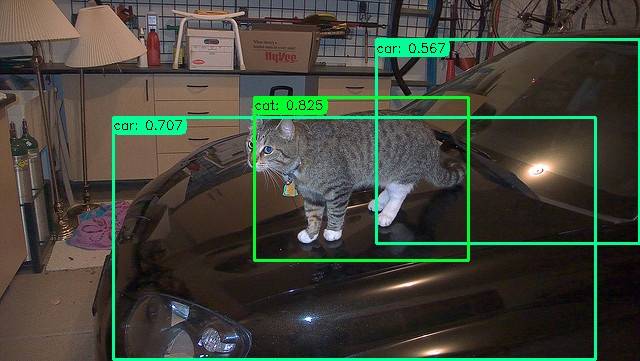}}
\end{minipage}
\hfill
\begin{minipage}[t]{0.195\linewidth}
  \centering
  \centerline{\includegraphics[width=3.5cm]{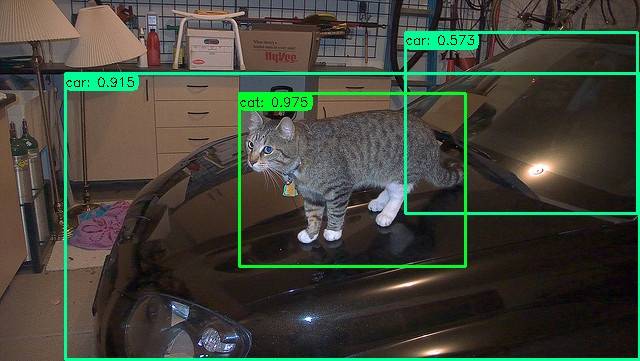}}
\end{minipage}
\hfill
\begin{minipage}[t]{0.195\linewidth}
  \centering
  \centerline{\includegraphics[width=3.5cm]{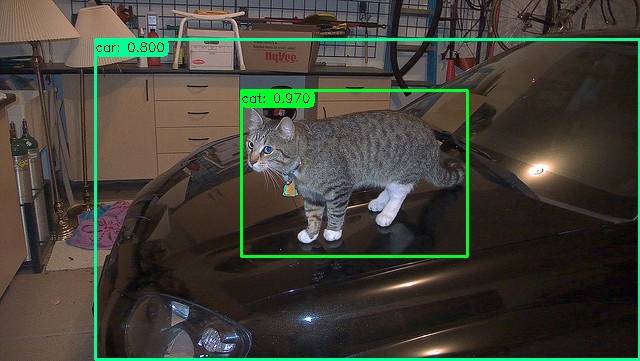}}
\end{minipage}\\
\begin{minipage}[t]{0.195\linewidth}
  \centering
  \centerline{\includegraphics[width=3.5cm]{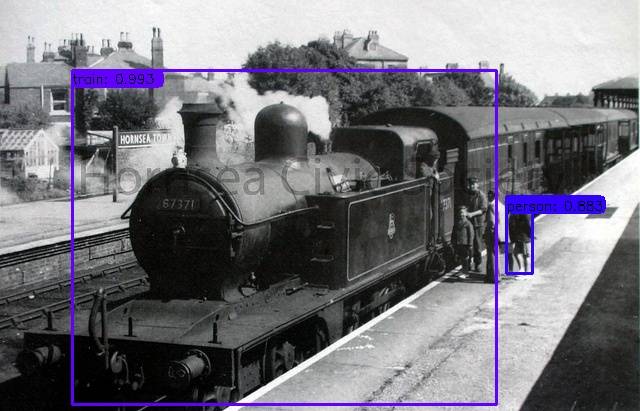}}
\end{minipage}
\hfill
\begin{minipage}[t]{0.195\linewidth}
  \centering
  \centerline{\includegraphics[width=3.5cm]{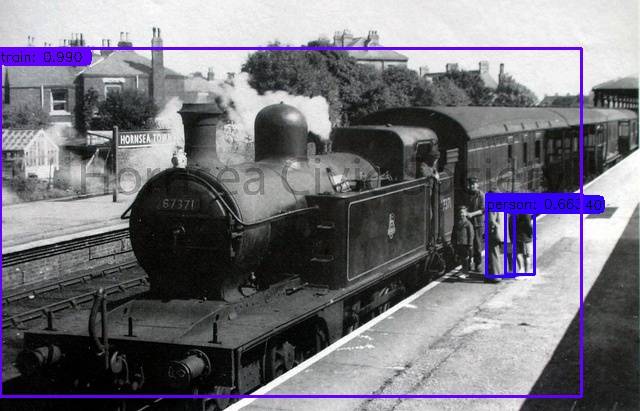}}
\end{minipage}
\hfill
\begin{minipage}[t]{0.195\linewidth}
  \centering
  \centerline{\includegraphics[width=3.5cm]{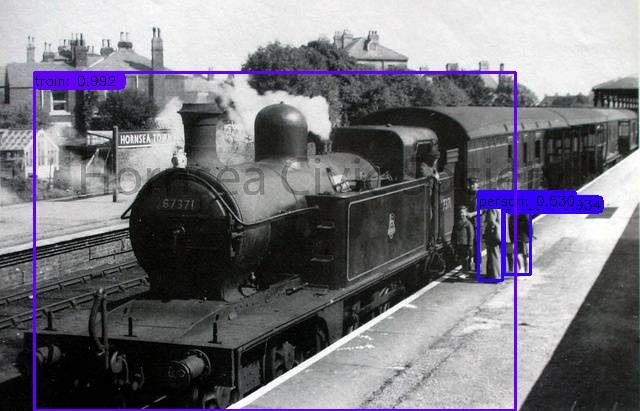}}
\end{minipage}
\hfill
\begin{minipage}[t]{0.195\linewidth}
  \centering
  \centerline{\includegraphics[width=3.5cm]{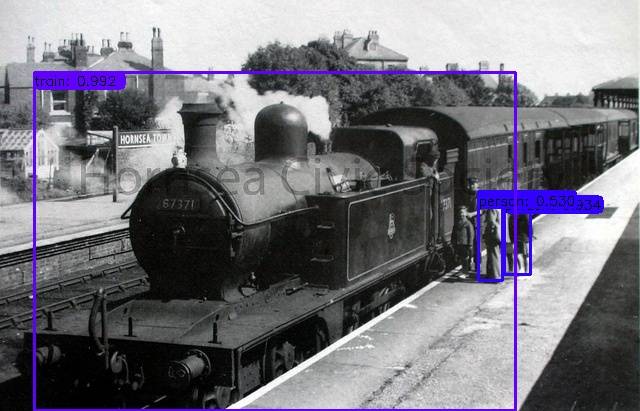}}
\end{minipage}
\hfill
\begin{minipage}[t]{0.195\linewidth}
  \centering
  \centerline{\includegraphics[width=3.5cm]{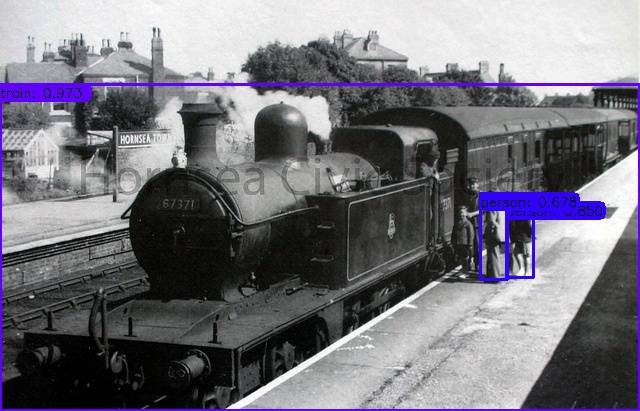}}
\end{minipage}\\
\begin{minipage}[t]{0.195\linewidth}
  \centering
  \centerline{\includegraphics[width=3.5cm]{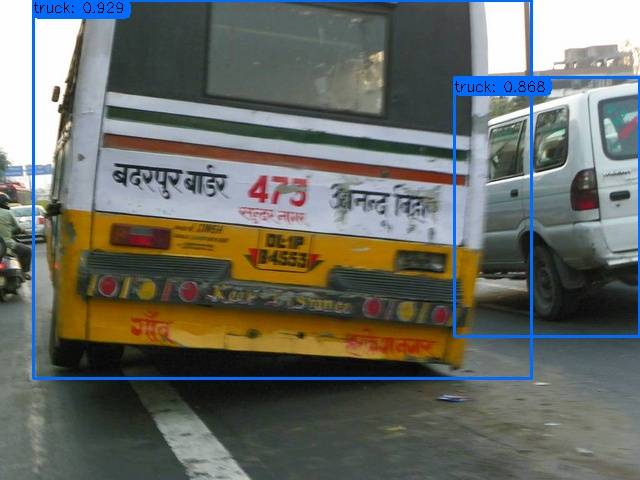}}
\end{minipage}
\hfill
\begin{minipage}[t]{0.195\linewidth}
  \centering
  \centerline{\includegraphics[width=3.5cm]{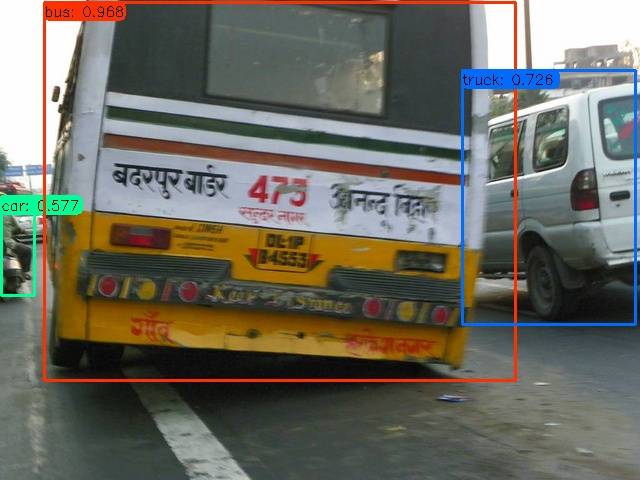}}
\end{minipage}
\hfill
\begin{minipage}[t]{0.195\linewidth}
  \centering
  \centerline{\includegraphics[width=3.5cm]{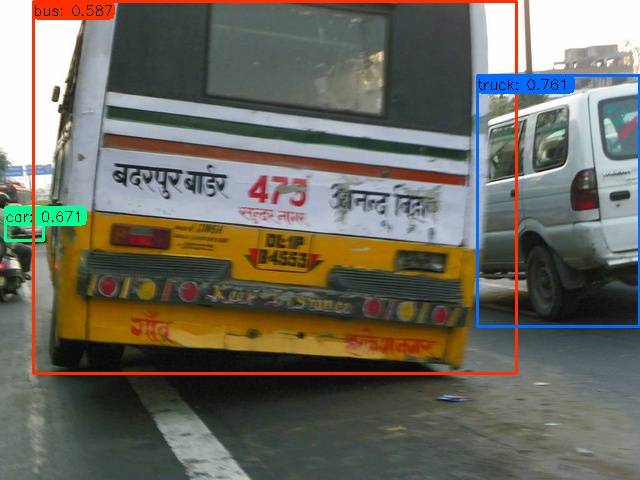}}
\end{minipage}
\hfill
\begin{minipage}[t]{0.195\linewidth}
  \centering
  \centerline{\includegraphics[width=3.5cm]{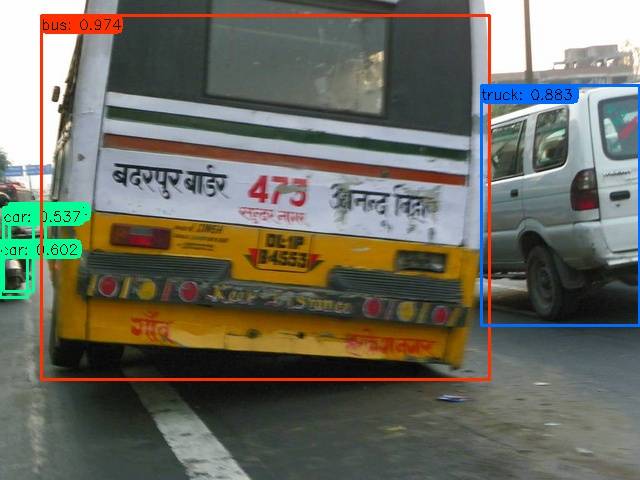}}
\end{minipage}
\hfill
\begin{minipage}[t]{0.195\linewidth}
  \centering
  \centerline{\includegraphics[width=3.5cm]{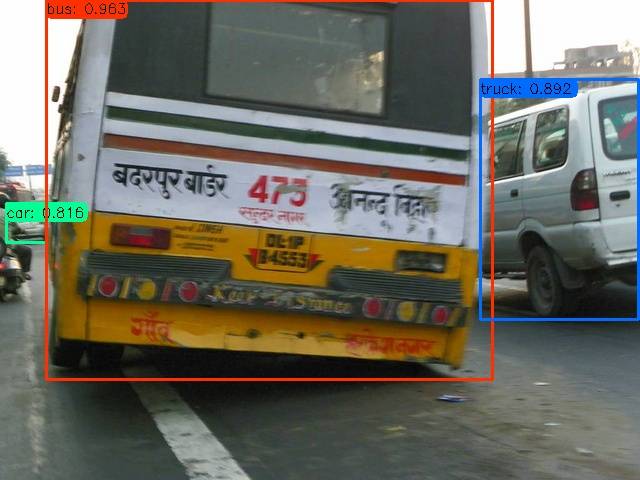}}
\end{minipage}\\
\begin{minipage}[t]{0.195\linewidth}
  \centering
  \centerline{\includegraphics[width=3.5cm]{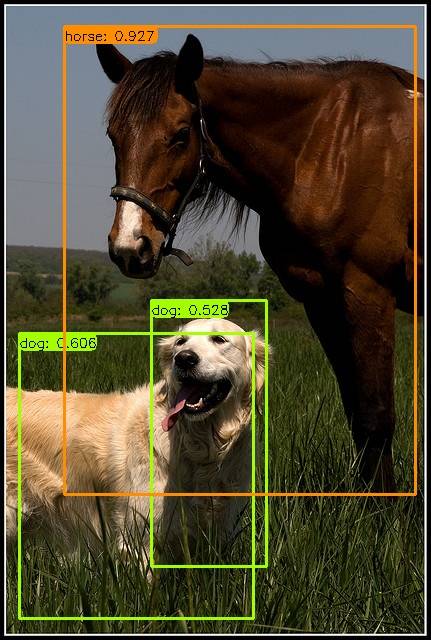}}
  \centerline{\small{(a) SSDLite+MobileNet}}
\end{minipage}
\hfill
\begin{minipage}[t]{0.195\linewidth}
  \centering
  \centerline{\includegraphics[width=3.5cm]{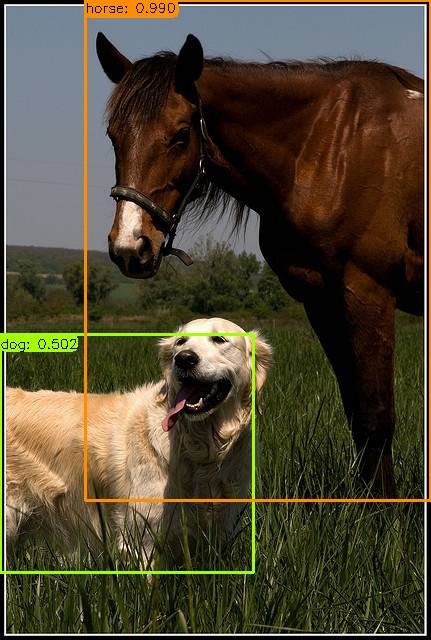}}
  \centerline{\small{(b) RefineDet+ShuffleNetV2}}
\end{minipage}
\hfill
\begin{minipage}[t]{0.195\linewidth}
  \centering
  \centerline{\includegraphics[width=3.5cm]{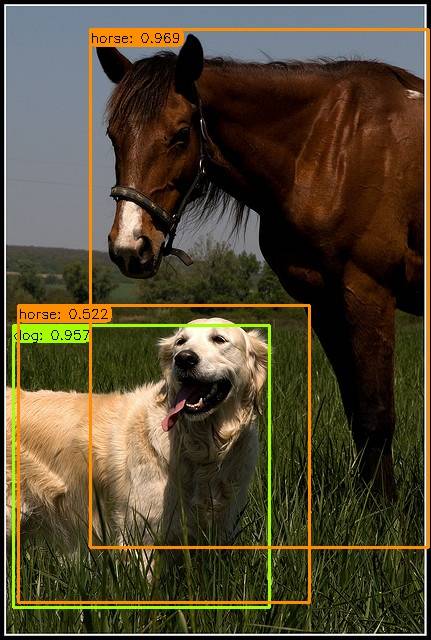}}
  \centerline{\small{(c) SSD+VGG}}
\end{minipage}
\hfill
\begin{minipage}[t]{0.195\linewidth}
  \centering
  \centerline{\includegraphics[width=3.5cm]{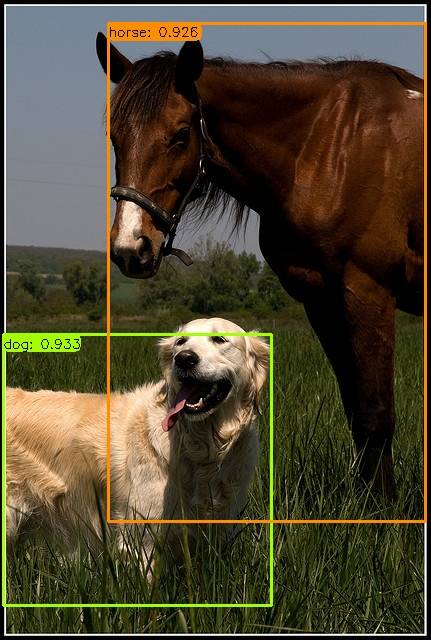}}
  \centerline{\small{(d) RefineDetLite}}
\end{minipage}
\hfill
\begin{minipage}[t]{0.195\linewidth}
  \centering
  \centerline{\includegraphics[width=3.5cm]{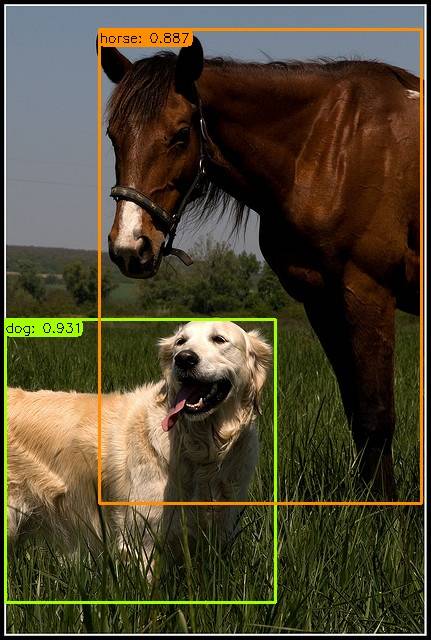}}
  \centerline{(e) RefineDetLite++}
\end{minipage}
\caption{Visualization comparisons among different models. This groups of images show that RefinDetLite++ can localize the prediction bounding boxes more accurately as compared to RefineDetLite. Best viewed by zooming in.}
\label{fig:visualization_accurate}
\end{figure*}

In this section, we provide more visualization cases (Fig.~\ref{fig:visualization_smallobjects}--\ref{fig:visualization_accurate}) of comparisons among different object detection algorithms, including SSDLite+MobileNet, SSD+VGG, RefineDet+ShuffleNetV2 and our proposed RefineDetLite and RefineDetLite++.

First, as demonstrated in Fig.~\ref{fig:visualization_smallobjects}, due to the the strong refining process and feature representation ability, our proposed RefineDetLite can detect more small objects (\eg, \textit{hot dogs, carrots, persons, oranges, chairs} and \textit{bicycles} from the first to the last row in turn). Furthermore, after adding the proposed training strategies, RefineDetLite++ can achieve better accuracy and recall.

Additionally, as shown in Fig.~\ref{fig:visualization_moreobjectsclasses}, RefineDetLite can also detect more kinds of objects, especially small objects, such as \textit{skateboards, chairs, skis, fir hydrants, bottles, traffic lights} and \textit{apples}. RefineDetLite can further refine the detection objects by revising some errors such as \textit{sports ball} in the first row and \textit{bird} in the last row.

Thanks to the IOU-guided loss and balanced multi-task training strategies, as compared to RefineDetLite, RefineDetLite++ can locate the objects more accurately. Fig.~\ref{fig:visualization_accurate} illustrated the comparisons in detail. The visualization results clearly show that RefineDetLite++ can achieve higher location accuracy such as \textit{motorcycle, umbrella, cake, car, train, bus, horse} and \textit{dog} in each row or images. Therefore, the $AP_{75}$ value of RefineDetLite++ ($31.0$) is much higher than the other lightweight detectors.

\end{document}